\documentclass[10pt,twocolumn,letterpaper]{article}

\PassOptionsToPackage{table}{xcolor}
\usepackage[pagenumbers]{cvpr} 

\usepackage{graphicx}
\usepackage{amssymb}
\usepackage{tikz}
\usepackage{array}
\usepackage{pifont}
\usepackage{geometry}
\usepackage{mathrsfs}
\usepackage{csquotes}
\usepackage{listings}
\usepackage{multirow}
\usepackage{adjustbox}

\definecolor{Best}{HTML}{C7EFCF}     
\definecolor{Second}{HTML}{FFE8A3}   
\definecolor{NAcol}{HTML}{D8D8D8}    

\newcommand{\good}[1]{\cellcolor{Best}{#1}}

\newcommand{\best}[1]{\cellcolor{Best}\textbf{#1}}
\newcommand{\secondbest}[1]{\cellcolor{Second}\underline{#1}}
\newcommand{\na}{\cellcolor{NAcol}--}
\newcommand{\colorsquare}[1]{\textcolor{#1}{\rule[0.5ex]{1.1ex}{1.1ex}}}

\usepackage{dblfloatfix} 
\usetikzlibrary{arrows.meta,positioning,calc,fit, decorations.pathreplacing}

\definecolor{codeblue}{rgb}{0.0,0.7,0.9}
\definecolor{codegray}{rgb}{0.5,0.5,0.5}
\definecolor{backcolour}{rgb}{0.98,0.98,0.98}

\newcommand{\cmark}{{\color{green!60!black}\ding{51}}}
\newcommand{\xmark}{{\color{red}\ding{55}}}
\newcommand{\simple}{{\color{magenta}Simple}}
\newcommand{\complex}{{\color{green!50!black}Cmplx.}}

\usepackage[most]{tcolorbox}
\usepackage{needspace}

\newtcolorbox{promptbox}[1][]{
  enhanced,
  colback=white,
  colframe=gray!60!black,
  boxrule=0.6pt,
  arc=2mm,
  left=2mm,
  right=2mm,
  top=1.2mm,
  bottom=1.2mm,
  title={\textbf{\textcolor{white!80!black}{#1}}},
  fonttitle=\bfseries\small,
}

\lstdefinestyle{CStyle}{
    language=C,
    backgroundcolor=\color{backcolour},
    basicstyle=\ttfamily\small,
    keywordstyle=\color{codeblue}\bfseries,
    commentstyle=\color{codegray}\itshape,
    stringstyle=\color{red!60!brown},
    numbers=none,
    showstringspaces=false,
    tabsize=4,
    breaklines=true,
    frame=single,
    rulecolor=\color{black!30},
    captionpos=b
}

\definecolor{cvprblue}{rgb}{0.21,0.49,0.74}
\usepackage[pagebackref,breaklinks,colorlinks,allcolors=cvprblue]{hyperref}

\usepackage{pifont}
\usepackage{enumitem}  
\usepackage{tikz}
\usepackage{amsmath}
\usetikzlibrary{3d,calc}
\usetikzlibrary{shapes.geometric, arrows.meta, calc}

\def\paperID{} 
\def\confName{CVPR}
\def\confYear{2026}

\title{A2Z-10M+: Geometric Deep Learning with A-to-Z BRep Annotations for AI-Assisted CAD Modeling and Reverse Engineering}

\vspace{-0.25cm}
\author{
\begin{tabular}{c c c c c}
{\textcolor{black}{Pritham K. Jena$^{1,2}$}} &
{\textcolor{black}{Bhavika Baburaj$^{1,2}$}} &
{\textcolor{black}{Tushar Anand$^{2}$}} &
{\textcolor{black}{Vedant Dutta$^{2}$}} & 
{\textcolor{black}{Vineeth Ulavala$^{2}$}}\\
\end{tabular} \\
\begin{tabular}{c}
\href{mailto:skaziz.ali@hyderabad.bits-pilani.ac.in}{\textcolor{black}{Sk Aziz Ali$^{1,2\dagger}$}}\\
\end{tabular} \\ [0.2cm]
\begin{tabular}{c c}
$^{1}$\text{\href{https://3dvision-lab.in/}{\textcolor{black}{3D Vision Group (3DVG)}}} & 
$^2$\text{\href{https://www.bits-pilani.ac.in/hyderabad/}{\textcolor{black}{BITS Pilani, Hyderabad, India}}} \\
\end{tabular} \\
\begin{tabular}{c c}
{\tt\small $^{\dagger}$Corresponding Author:} & {\tt\small \href{mailto:skaziz.ali@hyderabad.bits-pilani.ac.in}{\textcolor{black}{skaziz.ali@hyderabad.bits-pilani.ac.in}}}
\end{tabular}
}
\vspace{-0.7cm}
\begin{document}
\twocolumn[{%
\renewcommand\twocolumn[1][]{#1}%
\maketitle
\begin{center}
    \centering
    \captionsetup{type=figure}
    \includegraphics[width=1\linewidth]{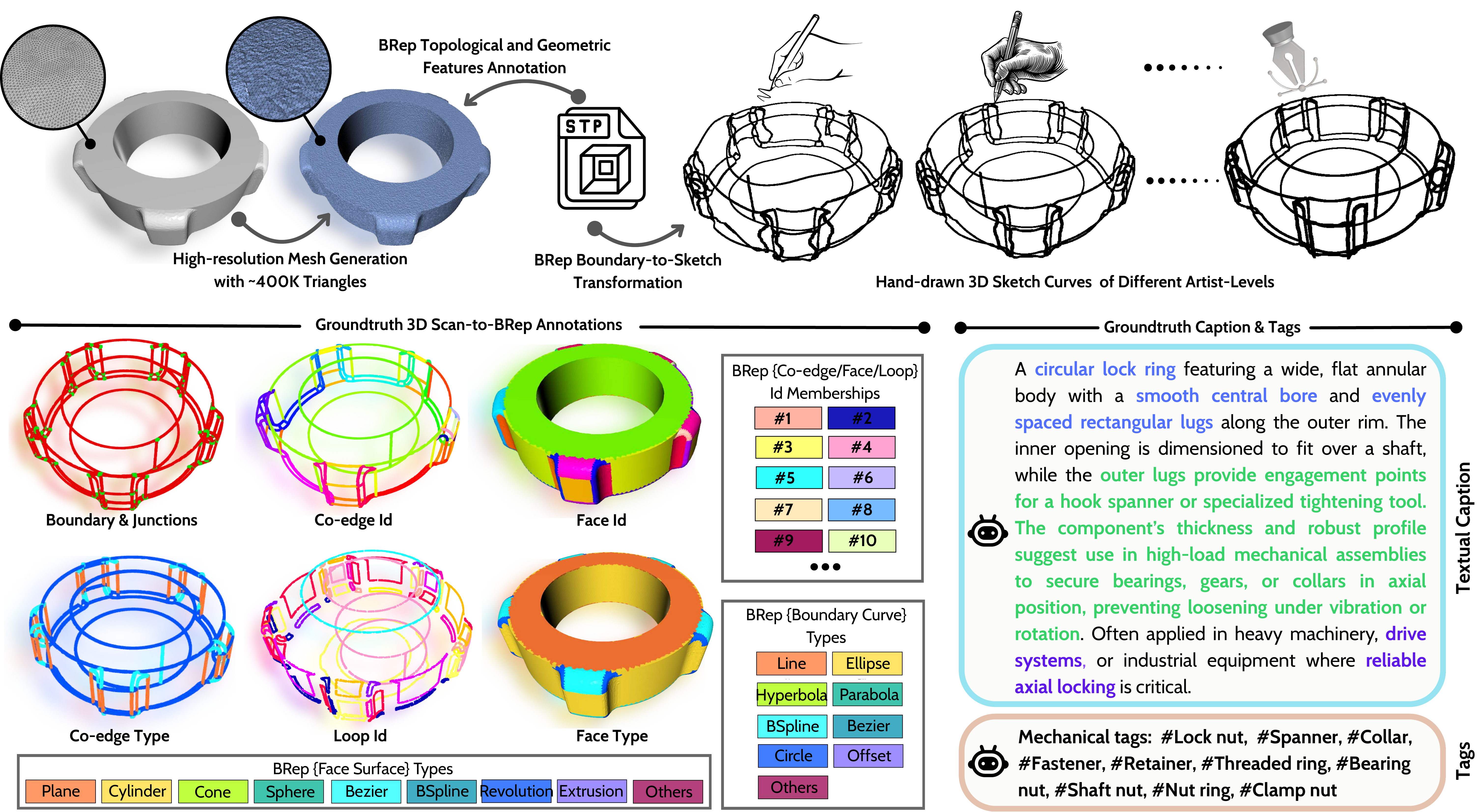}
    \captionof{figure}{\textbf{A2Z} data and annotations for multi-modal BRep learning in CAD reverse engineering
    (best viewed when zoomed in).}
   \label{fig:teaser}
\end{center}%
}]

\makeatletter
\let\SavedAddContentsLine\addcontentsline
\renewcommand{\addcontentsline}[3]{}%
\makeatother

\maketitle
\begin{abstract}
Reverse engineering and rapid prototyping of computer-aided design (CAD) models from 3D scans, sketches, or simple text prompts is vital in industrial product design. However, recent advances in geometric deep learning techniques lack a multi-modal understanding of parametric CAD features stored in their boundary representation (BRep). This study presents the largest compilation of 10 million multi-modal annotations for 1 million ABC CAD models, namely \textbf{A2Z}, to unlock an unprecedented level of BRep learning. A2Z comprises \textbf{(i)} high-resolution meshes with salient 3D scanning features, \textbf{(ii)} 3D hand-drawn sketches equipped with \textbf{(iii)} geometric and topological information about BRep co-edges, corners, and surfaces, and \textbf{(iv)} textual captions and tags describing the product in the mechanical world. Creating such carefully structured large-scale data — nearly 5 terabytes of storage — to leverage unparalleled CAD learning/retrieval tasks is very challenging. The scale, quality, and diversity of our multi-modal annotations are assessed using novel metrics, GPT-5, Gemini, and extensive human feedback mechanisms. To this end, we also merge an additional 25K CAD models of electronic enclosures (\textit{e.g.,} tablets, ports) created by skilled designers with our A2Z dataset. Subsequently, we train and benchmark a foundation model on a subset of 300K CAD models to detect BRep co-edges and corner vertices from 3D scans, a key downstream task in CAD reverse engineering. Project page and dataset details are available \text{\href{https://saali14.github.io/a2z.github.io/}{\textcolor{magenta}{here}}}. 
\end{abstract}
\vspace{-0.5cm}

%
%
%
%
%
%
%
%
%
%
%
%
%
%
\section{Introduction}\label{sec:intro}
\vspace{-0.1cm}
Geometry processing \cite{anderson2001discourseGeometryProc}, topological exploration \cite{weiler1986topological}, and understanding of design principles \cite{designIntent2018} are essential components of industrial product design \cite{cherng1998feature,REGASSAHUNDE2022100478} and reverse engineering \cite{BRepDetNet_2024ICMI,uy2022point2cyl}. 
In industrial product design and manufacturing, boundary representation (BRep) is the \textit{de facto} standard for feature-based geometry processing and topological exploration of CAD models \cite{lambourne2021BRepnet}.
As shown in the \cref{fig:teaser}, a BRep can be seen as a hierarchical structure of geometric primitives where solid shells of a CAD model are composed of surface primitives (\textit{e.g.,} plane, cylinder, cone), each bounded by a closed loop of one or many boundary curves called co-edges (\textit{e.g.,} line, circle, Bspline). Similarly, the co-edges are further connected by junction/corner vertices. These geometric primitives are connected components, and therefore, can be traversed through a topological walk \cite{weiler1986topological} (\textit{i.e.,} shell $\boldsymbol{\rightarrow}
$ faces $\boldsymbol{\rightarrow}
$ closed loops $\boldsymbol{\rightarrow}
$ co-edges $\boldsymbol{\rightarrow}
$ junctions) over the BRep complex chain \cite{guo2022complexgen}.

\vspace{0.1cm}
While the final version of any CAD model is stored in its BRep structure, its feature-based modeling includes sequential and repetitive design steps -- \textit{i.e.,} continuously drawing 2D sketches (\textit{e.g.,} parametric lines, arcs, or other curves) followed by a 3D operation (\textit{e.g.,} extrusion, revolution, fillet, etc.) \cite{DeepCAD_2021_ICCV}. These sequential design steps are collectively referred to as the \textit{design history}.
In recent times, the most popular supervised deep learning methods for CAD reverse engineering require high-quality 3D scans paired either with their \textit{design history} \cite{mkhan2024cadsignet, uy2022point2cyl, DeepCAD_2021_ICCV, multicad} or precisely annotated BRep information \cite{Point2CAD_Liu2024_CVPR, guo2022complexgen, sharma2020parsenet, BRepDetNet_2024ICMI, SpelsNet_NeurIPS24, Point2Primitive2025} for robust training. 

\vspace{0.1cm}
Obtaining the design history of all CAD models under ABC \cite{koch2019abc} requires proprietary access to the OnShape \cite{onshape} repository, which is non-permissive. At the same time, generating high-quality 3D scans from \textit{one million} low-poly meshes of ABC has never been done before\footnote{traditional NeRF is not an alternative to the 3D scanning process and we cannot manufacture \cite{kimmel2025position} a NeRF.}. Therefore, all \textit{design history}-based CAD construction methods depend upon $\sim$170K CAD samples from the DeepCAD \cite{DeepCAD_2021_ICCV} dataset. However, this dataset primarily comprises rudimentary CAD models, with the majority of shapes being \textit{simple} cuboid type, exhibiting minimal diversity in terms of geometry, topology, and shape categories. The main reason behind this is the selection of only those CAD models that are designed through simple extrusion operations on easily drawn 2D sketches. Furthermore, the input point clouds employed by methods like MultiCAD \cite{multicad}, DeepCAD \cite{DeepCAD_2021_ICCV}, CADSIGNet \cite{mkhan2024cadsignet}, and Point2Cyl \cite{uy2022point2cyl} consist of randomly subsampled points from the clean BRep face patches of CAD models. This is because of the unavailability of real scans as highlighted in \cite{mallis2023sharp}. In summary, this is a deadlock situation for design history-based CAD modeling methods to advance. The other group of BRep-based CAD reconstruction methods from input point clouds \cite{Point2CAD_Liu2024_CVPR, guo2022complexgen, sharma2020parsenet, BRepDetNet_2024ICMI, SpelsNet_NeurIPS24} uses either the Fusion-360 Gallery data \cite{willis2021fusion} with 8K samples or the ABCParts \cite{sharma2020parsenet} dataset, which is a small subset of just 32K samples from ABC \cite{koch2019abc}. Though the BRep information is available in the \lq\textit{.step}\rq ~file of a CAD model, the corresponding high-quality 3D scans and annotations are not available.  
%
%
\renewcommand{\arraystretch}{1}
\begin{table}[t]
\centering
\scriptsize
\renewcommand{\arraystretch}{1.2}
\begin{tabular}{|>{\centering\arraybackslash}m{1.6cm}|
                >{\centering\arraybackslash}m{1.0cm}|
                >{\centering\arraybackslash}m{1.0cm}|
                >{\centering\arraybackslash}m{0.2cm}|
                >{\centering\arraybackslash}m{0.2cm}|
                >{\centering\arraybackslash}m{0.2cm}|
                >{\centering\arraybackslash}m{0.2cm}|
                >{\centering\arraybackslash}m{0.2cm}|
    }
\hline
\parbox{1.6cm}{\centering \textbf{Dataset}} & 
\parbox{1.0cm}{\centering \textbf{CAD Models}} & 
\rotatebox{90}{\parbox{1.2cm}{\centering \textbf{\#Models}}} & 
\rotatebox{90}{\parbox{1.5cm}{\centering \textbf{BRep Labels}}}& 
\rotatebox{90}{\parbox{1.5cm}{\centering \textbf{Text Caption}}} & 
\rotatebox{90}{\parbox{1.0cm}{\centering \textbf{3D Scans}}} & 
\rotatebox{90}{\parbox{1.1cm}{\centering \textbf{3D Sketch}}} & 
\rotatebox{90}{\parbox{1.6cm}{\centering \textbf{Design History}}}\\
\hline
DeepCAD \cite{DeepCAD_2021_ICCV} & \simple~\cmark & 170,000+ & \xmark  & \xmark & \xmark & \xmark & \cmark\\
\hline
Fusion360 \cite{willis2021fusion}     & \simple~\cmark  & $\sim$8,000 & \xmark  & \xmark & \xmark & \xmark & \cmark\\
\hline
CADParser \cite{zhoucadparser2023}    & \simple~\cmark  & 41,000 & \xmark  & \xmark & \xmark & \xmark & \cmark\\
\hline
CAD-MLLM  & \simple~\cmark & 453,220 & \xmark & \cmark & \xmark & \xmark & \cmark\\
\hline
ABC \cite{koch2019abc} & \complex~\cmark & 1,000,000  & \xmark & \xmark & \xmark & \xmark  & \xmark \\
\hline
CC3D-Ops \cite{CADOPsNetDV22} & \complex~\cmark & $\sim$50,000 & \cmark & \xmark & \cmark & \xmark & \xmark \\
\hline
ABCPrimitive \cite{huang2021primitivenet} & \simple~\cmark & 5,619 & \cmark & \xmark & \xmark & \xmark & \xmark \\
\hline
\textbf{A2Z (Ours)} &\complex~\cmark & 1,025,000+ & \cmark & \cmark & \cmark & \cmark & \xmark\\
\hline
\end{tabular}
\caption{Comparison of open-source CAD datasets w.r.t \textbf{A2Z}}
\label{tab:dataset-comparison}
\end{table}

\vspace{0.1cm}
The previously mentioned scarcity of large-scale data and the lack of rich multi-modal annotations remain major obstacles in advancing CAD reverse engineering research. These limitations also restrict research progress in BRep reconstruction, developing novel approaches like freehand 3D sketch-to-BRep generation, text-to-BRep modeling, and other broader semantic reasoning tasks for \textit{complex} CAD models. As a result, existing methods for BRep parsing and fitting -- like BRepSeg \cite{BRepSeg_2025_TIP}, ParseNet \cite{sharma2020parsenet}, HPNet \cite{yan2021hpnet}, SPFN \cite{spfn}, its extension CPFN \cite{le2021cpfn}, and ComplexGen \cite{guo2022complexgen} -- are not robust enough (reasons are highlighted in \cref{sec:2_relatedwork}). 
%
%
%
%
To overcome these data challenges and leverage many new avenues of multi-modal BRep learning, we derive \textit{A2Z} as the largest compilation of $>$10 million annotations for $>$1 million complex CAD models. 
%
%
%
%
In \cref{tab:dataset-comparison}, we elaborate on different open-access datasets and their multi-modal annotations compared to our \textbf{\textit{A2Z}}. This illustrates the present importance and further impact our proposed A2Z dataset can make in advancing supervised X-to-BRep representation \cite{jayaraman2021uvNet, lambourne2021BRepnet} learning. Our main \textbf{contributions} are:

\vspace{0.05cm}
\noindent\textbf{3D Scans with BRep Labels.} We process $1$ million low-poly CAD meshes of ABC and upscale them into 16x higher resolution meshes. Each vertex is labeled by its corresponding BRep co-edge, junction, and face information using a proximity-aware, multi-threshold assignment policy (see \cref{subsec:3_ProximAware_Annot}). We also propose a geometric transformation algorithm that imprints natural artifacts observed in real 3D scans \cite{cc3d, mallis2023sharp} -- \textit{e.g.,} surface dents, missing parts, protrusions near small holes, and surface roughness (see \cref{subsec:3_Scans_Artifacts}). The resulting \textbf{A2Z} is the largest dataset for scan-to-CAD reverse engineering.  

\vspace{0.1cm}
\noindent\textbf{Multi-Level 3D Sketching:} We generate 3D sketches using a novel algorithm that simulates different skill levels of drawing artists. Using normalized arc-length parameterization of BRep co-edge curves, we induce realistic deformation \cite{FlashArmMovement1688}, distortion, intentional endpoint openings, planar tapering \cite{WindowBow78}, and other haptic sketching traits \cite{SketchyRendering2012} (see \cref{subsec:3_HandDrawn_Sketching}).        

\vspace{0.1cm}
\noindent\textbf{Vision-Language Models (VLMs) for Captions.} A2Z is the first to automate high-quality captions ($\leq$\,200 words per model) and tags ($\leq$\,20 per model) for the entire ABC corpus using a cost-effective VLM system (see \cref{subsec:3_AnnotTextImages}). Inspired by ImageNet \cite{ImageNet} and WordNet \cite{fellbaum1998wordnet}, we also derive a hierarchical class and category structure (with a depth of up to 4) based on a key ontology.  

\vspace{0.1cm}
\noindent\textbf{Additional CAD of Electronic Enclosures.} We employ skilled designers to create 25K CAD models of electronic enclosures -- \textit{e.g.,} charging ports, thin casings, and sockets -- on OnShape \cite{onshape}. These models, with annotations, are merged into A2Z (see \cref{sec:3_ElectronicEnclosureData}).    

\vspace{0.1cm}
\noindent\textbf{Benchmarking A2Z.} Finally, we train and benchmark a geometric deep learning model for BRep boundary and junction detection tasks (see \cref{sec:4_GeometricDL_Models} and \cref{subsec:ModelEval}). Data analytics, human feedback scores of A2Z multi-modal annotations, and comparisons with Gemini \cite{gemini} and GPT-5 \cite{achiam2023gpt} are systematically staged (see \cref{subsec:DataEval}).

\vspace{-0.1cm}
\section{Related Work}
\label{sec:2_relatedwork}
\vspace{-0.1cm}
\subsection{Supervised Scan-to-BRep Learning}\label{subsec:2_Scan2BRep_Methods}
\vspace{-0.1cm}
Many recent methods for B-Rep learning \cite{guo2022complexgen, StitchAShape2025, BRepDetNet_2024ICMI, Point2CAD_Liu2024_CVPR} follow a common pipeline. They first estimate junction and boundary vertices from an input point cloud and learn edges-to-corners connectivity to form a topological graph, yielding a CAD wireframe. PIE-Net \cite{wang2020pie}, BRepDetNet \cite{BRepDetNet_2024ICMI}, and ComplexGen \cite{guo2022complexgen} focus on boundary and corner detection as the initial challenges in BRep learning. Conversely, ParseNet \cite{zhoucadparser2023}, BGPSeg \cite{BRepSeg_2025_TIP}, SPFN \cite{spfn}, and CPFN \cite{le2021cpfn} perform BRep surface segmentation as their primary task for CAD construction. Only Point2CAD \cite{Point2CAD_Liu2024_CVPR} and ComplexGen \cite{guo2022complexgen} advance to parametric surface fitting. However, estimating adjacency \cite{lee2025brepdiff, CADDreamer_2025_CVPR, DTBRepGen_CVPR25} through point membership assignment into boundaries, corners, loops, and faces is never jointly addressed, mainly due to the absence of exhaustive BRep labels in public datasets (see \cref{tab:dataset-comparison}).

\vspace{-0.1cm}
\subsection{Multi-modal Annotations for CAD}\label{subsec:2_MultiModal_Dataset}
\vspace{-0.1cm}
Recent years have seen a rapid rise in publicly available BRep datasets \cite{willis2021fusion, koch2019abc, cc3d, HoLABRep_2025, MFCAD2021, jayaraman2021uvNet}. However, except for CC3D \cite{cc3d, mallis2023sharp}, most datasets contain synthetic CAD models (in \lq.step\rq~file format) without paired 3D scans. For instance, MFCAD \cite{MFCAD2021}, FabWave \cite{Starly2019FabWaveCR}, SolidLetters \cite{jayaraman2021uvNet}, Fusion-360 \cite{willis2021fusion}, and ABC \cite{koch2019abc} are all synthetic CAD models designed on the Fusion-360 API \cite{Fusion360API} or Onshape \cite{onshape} platform. Although BRep information parsing from these datasets is feasible, there remains a lack of consistent class-level annotations, 2D/3D free-form sketches, realistic 3D scans, multi-view images (2D), and textual captions or tags (1D). For example, state-of-the-art generative BRep methods like SolidGen \cite{2022SolidGen} and BRepGen \cite{xu2024brepgen} use the DeepCAD dataset and partially selected ($<$90K) CAD models of ABC. Meanwhile, other BRep modeling methods exploiting synthetically rounded boundaries \cite{CADGenRoundedVox2022}, smooth fillets \cite{jiang2025defillet}, or sharp boundaries \cite{matveev2022def} rely on varied workarounds across datasets. Although many datasets exist for 3D content generation from text \cite{li2023generative}, sequential CAD design steps from sketches \cite{Sketch2CAD, li2022free2cad}, or text inputs \cite{khan2024text2cad}, none provide large-scale, unified, and multi-modal annotations. Only HoLABRep \cite{HoLABRep_2025} has aligned ABCPrimitive \cite{huang2021primitivenet} with textual and basic sketch annotations.

\vspace{-0.2cm}
\section{A2Z Data and Annotation Processing}
\label{sec:A2Z_Dataset}
\vspace{-0.1cm}
We treat a boundary representation (BRep) \(\boldsymbol{\mathcal{B}}\) of a CAD model as a chain complex $\boldsymbol{\mathcal{C}} = (\boldsymbol{\mathcal{V}}, \boldsymbol{\mathcal{E}}, \boldsymbol{\mathcal{F}}, \boldsymbol{\mathcal{A}})$ where the set of faces $\boldsymbol{\mathcal{F}} \in \mathbb{R}^{n_f \times 1}$ is defined by their ids and stores surface parameters along with other features, co-edges $\boldsymbol{\mathcal{E}} \in \mathbb{R}^{n_e \times 1}$ are defined by their ids and store likewise parametric curve features, and corner vertices $\boldsymbol{\mathcal{V}} \in \mathbb{R}^{n_v \times 1}$ are defined by their ids along with Euclidean coordinates (see \cref{fig:teaser}). Therefore, the 2D adjacency matrices $\boldsymbol{EV}\in\{0,1\}^{n_e \times n_v}$ and $\boldsymbol{FE}\in\{0,1\}^{n_f\times n_e}$ can determine the edges-to-corners and faces-to-edges topological connectivity. In addition, the topological walk along the co-edges is also expressed by the matrix $\boldsymbol{\mathcal{A}}$ as a tuple of three transition vectors $(\boldsymbol{\mathcal{N}}, \boldsymbol{\mathcal{P}}, \boldsymbol{\mathcal{M}})\in\mathbb{R}^{n_e \times 1} $, namely next, parent, and mate. For instance, to know the \textit{the mating face id of another face that is next to my current co-edge id $e$}, we can express the topological walk as $\boldsymbol{\mathcal{F}}\left[
\boldsymbol{\mathcal{M}}\left[ \boldsymbol{\mathcal{N}}\left[\boldsymbol{\mathcal{E}}\left[e\right]\right]\right]\right]$ (see \cite{jayaraman2021uvNet}). During the annotation process, we transfer the true IDs, topological connections, and geometric features of $\boldsymbol{\mathcal{B}}$ to every vertex $\mathbf{p}_i$ of its paired 3D scan $\mathcal{P}=\{\mathbf{p}_i\}_{i=1}^N$ that corresponds to either $\boldsymbol{\mathcal{V}}, \boldsymbol{\mathcal{E}}, \text{or } \boldsymbol{\mathcal{F}}$. BRep primitives like loops $\boldsymbol{\mathcal{L}}$ can also be derived as the chord-free cycles of edges $\in\boldsymbol{\mathcal{E}}$, or shells can be as a closed volume of faces $\in\boldsymbol{\mathcal{F}}$.       
\vspace{-0.1cm}
\subsection{High Resolution Meshes like 3D Scans}\label{subsec:3_Scans_Artifacts}
We begin with a low-polygon triangular mesh 
\(\mathcal{M}^{\text{init}} = (\mathcal{V}^{\text{init}}, \mathcal{T}^{\text{init}})\) 
sourced from the ABC dataset, together with its corresponding BRep \(\boldsymbol{\mathcal{B}}\). 
Our main objective is to transform \(\mathcal{M}^{\text{init}}\) into a high-resolution, 
scan-like mesh \(\mathcal{M}\) that accurately reflects the imperfections, occlusions, 
noisy surface perturbations, protrusions, and cavities commonly observed in data acquired by short-range handheld scanners, \textit{e.g.,} Artec3D (\url{artec3d.com}) or Scantech devices. These transformations are carried out through a sequence of geometry processing steps.
\begin{figure}[t]
  \centering
    \includegraphics[width=0.99\linewidth]{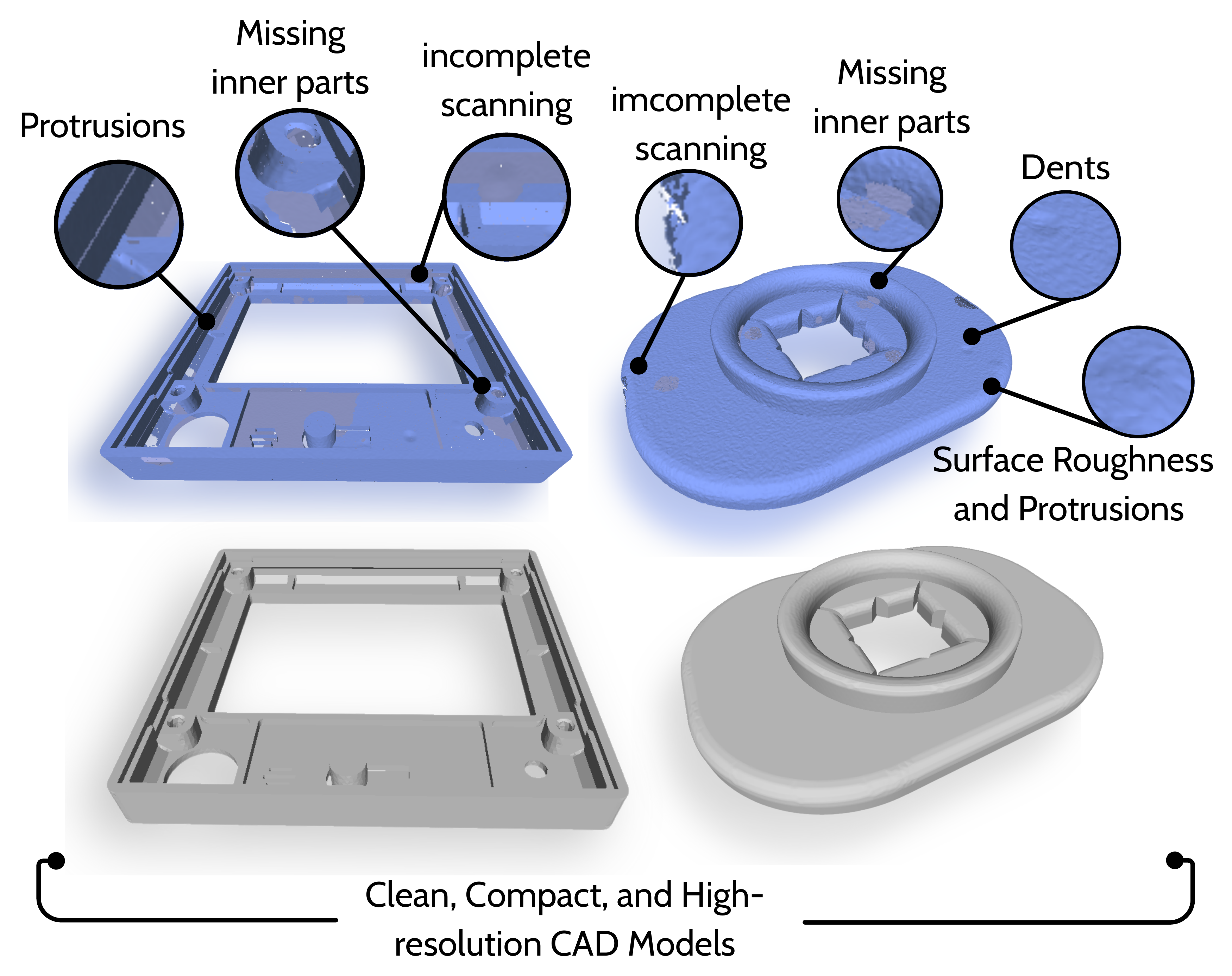}
   \caption{Low-poly CAD meshes are compared to high-quality scans after a series of geometry processing steps (\cref{subsec:3_Scans_Artifacts}).}
   \label{fig:CAD2Scans}
\end{figure}
\vspace{-0.3cm}
\paragraph{Step-I: Mesh Upsampling.}  
We apply two successive iterations of midpoint subdivision of triangles in $\mathcal{M}^{init}$.  
This results in a refined mesh 
\(\mathcal{M} = (\mathcal{V}, \mathcal{T})\) 
with approximately \(1.5 \times 10^{5}\) vertices and \(3.8 \times 10^{5}\) triangles, 
a density comparable to meshes reconstructed from high-precision scanning systems. Therefore, the vertices $\mathcal{V}$ are now equivalent to $\mathcal{P}$. To apply the transformations, we dense sample vertices $\boldsymbol{x}$ on the parametric uv-grid \cite{jayaraman2021uvNet} of $\boldsymbol{\mathcal{F}}$, arc-length  $\boldsymbol{\mathcal{E}}$, and corners $\boldsymbol{\mathcal{V}}$ of $\boldsymbol{\mathcal{B}}$.

\vspace{-0.3cm}
\paragraph{Step-II: Tangential Shrinkage Near Small Loops.}  
To replicate realistic scan imperfections, we first identify small loops in $\boldsymbol{\mathcal{B}}$.  
For each loop \(\ell \in \boldsymbol{\mathcal{L}}\), its perimeter \(L_{\ell}\) is compared against the maximum \(L_{\max}=\max_{\ell \in \mathcal{L}} L_{\ell}\). The loops with \(L_{\ell}/L_{\max}<\tau_h\) (given \(|\boldsymbol{\mathcal{L}}|>2\)) are classified as tiny holes \(\mathcal{H}\).  
Such small openings are prone to poor capture due to sensor frustum limitations (see \cref{fig:CAD2Scans}).  
To mimic this effect, vertices adjacent to \(\ell \in \mathcal{H}\) are shrunk tangentially toward the loop centroid \(\boldsymbol{c}_{\ell}\) using the in-plane projector \(P_{\ell}=I-\boldsymbol{n}_{\ell}\boldsymbol{n}_{\ell}^{\!\top}\):
\[
\vspace{-0.2cm}
\Delta\boldsymbol{x}_{\text{shrink}}=\boldsymbol{x}-\eta_{\ell}\,w_{\ell}(\boldsymbol{x})\,P_{\ell}(\boldsymbol{x}-\boldsymbol{c}_{\ell}),
\]  
thereby introducing protrusions and small gaps that simulate visibility loss.  
In addition, incomplete coverage is reinforced by removing a controlled subset of triangles from the mate-face \(f_{\text{mate}}(\ell)\in\boldsymbol{\mathcal{F}}\), subject to  
\[
\vspace{-0.2cm}
\sum_{t\in\mathcal{T}_{\text{miss}}(\ell)} \operatorname{Area}(t) \le 0.2\,\operatorname{Area}(f_{\text{mate}}(\ell)),
\] which models the occlusion of regions lying behind narrow openings.  

\vspace{-0.4cm}
\paragraph{Step-III: Adding Surface Roughness.}  
Real-world scanned surfaces exhibit both measurement error and 
small irregularities.  
We emulate this by displacing each vertex $\boldsymbol{x}$ along its normal $\boldsymbol{n}(\boldsymbol{x})$ using a multi-octave Perlin noise \cite{Perlin1985} field 
\vspace{-0.3cm}
\(N : \mathbb{R}^3 \to [-1, 1]\):
\[
\vspace{-0.2cm}
\Delta\boldsymbol{x}_{\text{rough}} =
A_r \sum_{o=1}^O \lambda_o \, N(\omega_o \boldsymbol{x}) \, \boldsymbol{n}(\boldsymbol{x}),
\]
with \(\sum_o \lambda_o = 1\) and \(\omega_o\) as increasing scalar field across octaves $O$.  
This step injects millimeter-scale inaccuracies while preserving sharp edges.

\vspace{-0.4cm}
\paragraph{Step-IV: Adding Dents and Protrusions.}  
In the final step, to capture manufacturing artifacts, we randomly select planar B-Rep faces 
and define \(K\) seed points \(\{\boldsymbol{s}_k\}_{i=1}^{K}\).  
Around each seed, we introduce localized circular dents and subtle bumps based on the outward or inward direction of the seed point's normal vector:
\vspace{-0.2cm}
\[
\begin{aligned}
\Delta\boldsymbol{x}_{\text{dent}|\text{bump}} &=\\
\sum_{k=1}^{K}\!\left(-\delta_k\,e^{-\frac{\|P_{f}(\boldsymbol{x}-\boldsymbol{s}_k)\|_2^{2}}{2\rho_k^{2}}}
+\varepsilon_k \sin\!\tfrac{\pi\|P_{f}(\boldsymbol{x}-\boldsymbol{s}_k)\|_2}{\rho_k}\right)
&\times\boldsymbol{n}(\boldsymbol{x}).
\end{aligned}
\] where \(P_f\) projects onto the face plane, \(\delta_k\) sets dent depth, 
\(\rho_k\) the affected radius, and \(\varepsilon_k\) the amplitude of protrusion or bump (see \cref{fig:CAD2Scans}).


%
%
\vspace{-0.1cm}
\subsection{Proximity-Aware Smooth Annotations}\label{subsec:3_ProximAware_Annot}
\begin{figure}[t]
  \centering
  \includegraphics[width=1.0\linewidth]{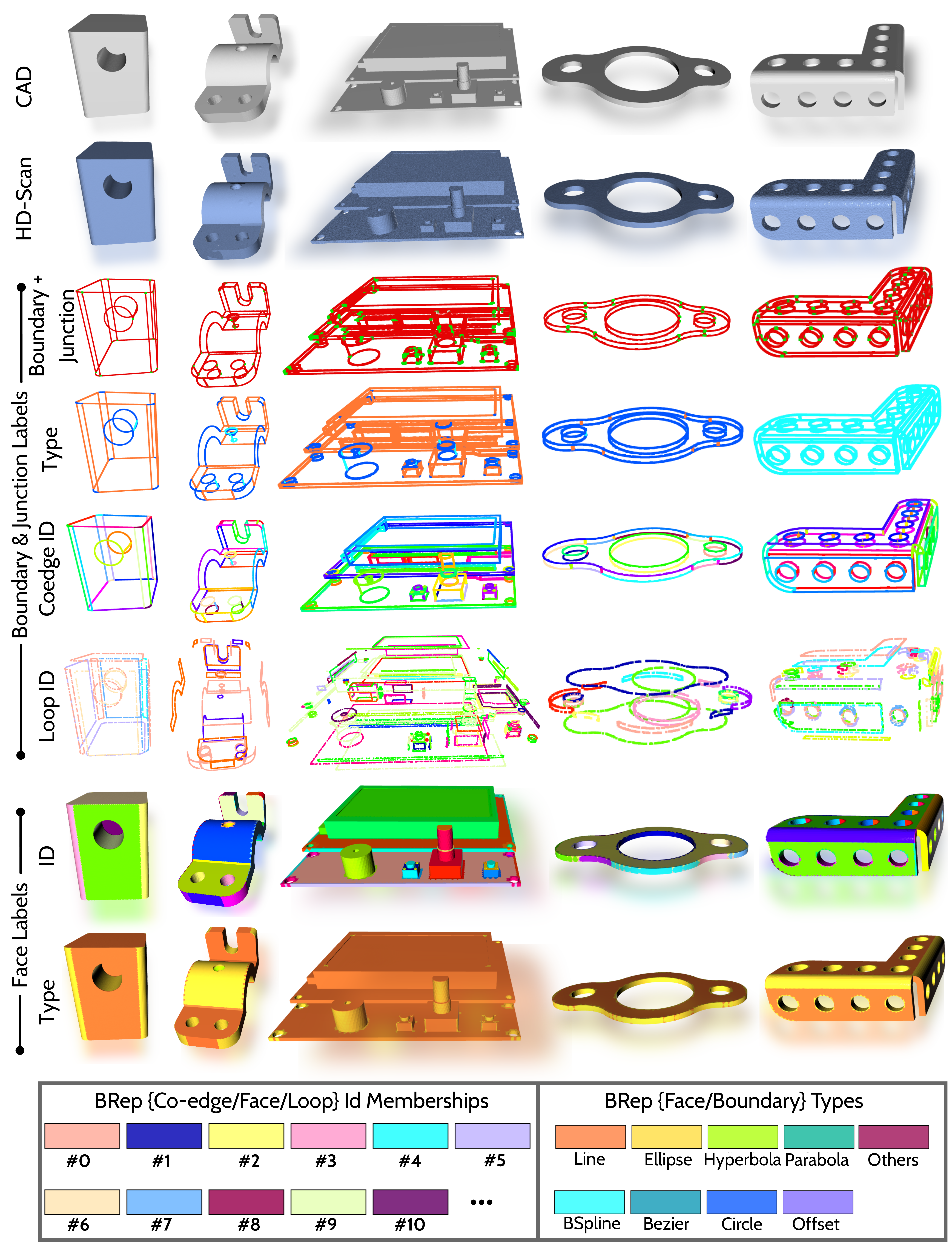}
   \vspace{-0.2cm}
   \caption{BRep annotations on 3D scans of \textbf{A2Z} dataset.}
   \label{fig:Annotation_Quality}
\end{figure}

We replace the most commonly practiced hard nearest–neighbor rule with a novel policy of multiscale, BRep coedge length–aware, and SPH–weighted \cite{SPHMoaghan92} membership assignment\footnote{Detailed formulations are in the supplement.} 
Given the 3D scan points \(\mathcal{P}=\{\boldsymbol{p}_i\}_{i=1}^{N}\) from the high-quality mesh, we use the sampled points \(\boldsymbol{x}\) on $\boldsymbol{\mathcal{B}}$ that have already gone through the transformation steps described in \cref{subsec:3_Scans_Artifacts}, i.e.,
\vspace{-0.2cm}
\begin{equation}
    \boldsymbol{x} \mapsto \boldsymbol{x} + 
    \Delta\boldsymbol{x}_{\text{shrink}} +
    \Delta\boldsymbol{x}_{\text{rough}} +    
    \Delta\boldsymbol{x}_{\text{dent}|\text{bump}}
    \vspace{-0.1cm}
\end{equation}
and define a local frame around $\boldsymbol{x}$. Next, the SPH weights $W_k^{\text{sph}}$ are aggregated across different neighborhoods $\mathcal{N}(\boldsymbol{x})$ of $\boldsymbol{x}$ at $K$ different scales as 
\vspace{-0.2cm}
\begin{equation}
\omega_i(\boldsymbol{x})= 
\sum_{k=1}^{K} w_k\,
W_k^{\text{sph}}(\boldsymbol{x}) 
\vspace{-0.1cm}
\end{equation}
and choose the probabilistic soft labels $\pi(\boldsymbol{x})$, that \emph{does not} miss nearby candidates, defined by
\[
\vspace{-0.2cm}
p_i(\boldsymbol{x})=\frac{\omega_i(\boldsymbol{x})}{\sum_{j\in\mathcal{N}(\boldsymbol{x})}\omega_j(\boldsymbol{x})},
\,\,
\pi(\boldsymbol{x})=\arg\max_{\,i\in\mathcal{N}(\boldsymbol{x})}p_i(\boldsymbol{x}),
\] for a given scan point $\boldsymbol{p}_i$.

\vspace{-0.3cm}
\paragraph{Boundary annotations:} Each vertex $\boldsymbol{p}_i$ that is labeled as a boundary now stores the parent edge id $e$ in \(\mathcal{E}\), its corresponding loop id $\ell$ in \(\boldsymbol{\mathcal{L}}\), the mate loop id $\boldsymbol{\mathcal{L}} \left[\boldsymbol{\mathcal{M}}\left[ \boldsymbol{\mathcal{E}}\left[e\right]
\right]\right]$, the two incident faces in \(\boldsymbol{\mathcal{F}}\), and a feature vector of the curve primitive (line, circle, spline). Extra scalar fields are reserved for future use.  
%
%
\paragraph{Junction annotations:} For every junction (\textit{i.e.,} coedge endpoint), we record the scan–vertex indices as per the soft-label assignment together with the ids of the adjacent loops, edges, and faces.  
\vspace{-0.5cm}
\paragraph{Face annotations:} Each scan vertex \(\boldsymbol{p}_i\) is assigned to the closest face in \(\boldsymbol{\mathcal{F}}\) by projection, along with a surface-class code (plane, cylinder, sphere, etc.).

\begin{figure}[t]
  \centering
    \includegraphics[width=1.1\linewidth]{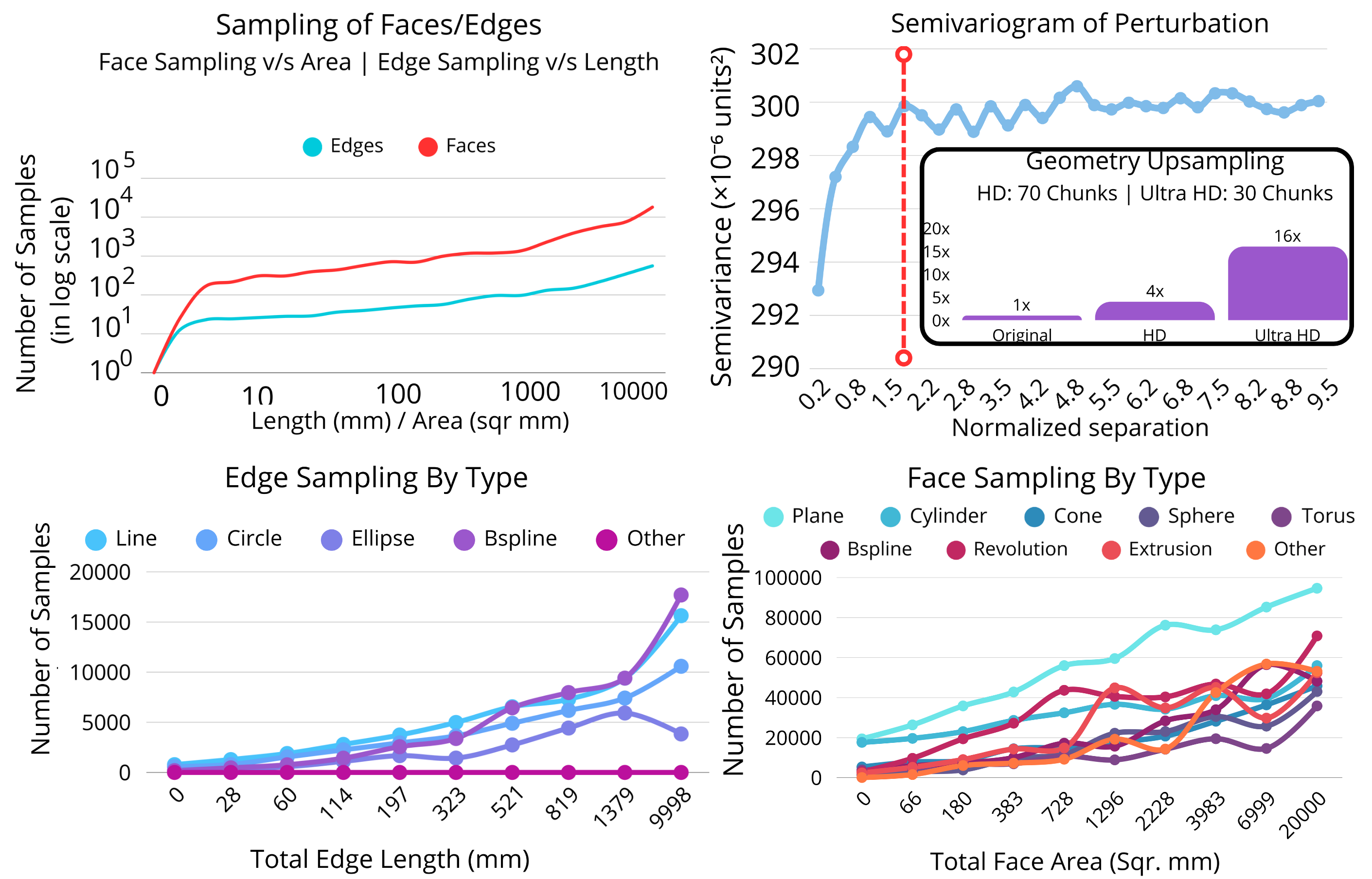}
   \caption{Data and annotation analytics of \textbf{A2Z}}
   \label{fig:AnnotAnalytics}
\end{figure}

\noindent\cref{fig:Annotation_Quality} shows how this concise annotation space fuses topology and geometry, offering rich supervision for deep learning tasks such as boundary detection, loop reasoning, edge typing, and face classification in Scan-to-BRep pipelines. The quantitative data analytics in \cref{fig:AnnotAnalytics} illustrate how the distribution of different numbers of CAD models varies when sampled across different areas, face types, co-edge lengths, and types of faces/edges. The top right semivariogram \cite{Semivariogram} plot in \cref{fig:AnnotAnalytics} explains the spatial structure of our high-quality scans $\mathcal{P}$ w.r.t a normalized separation distance between a group of points from any randomly selected centroid.    
\subsection{3D Hand-Drawn Sketches of BRep}\label{subsec:3_HandDrawn_Sketching}
To generate a 3D sketch from the co-edges $\boldsymbol{\mathcal{E}}\in\boldsymbol{\mathcal{B}}$, we apply three different types of displacement fields  \(\boldsymbol{\Delta \Vec{u}}_{\text{line}}(s), \boldsymbol{\Delta \Vec{u}}_{\text{arc}}(s), \boldsymbol{\Delta \Vec{u}}_{\text{gen}}(s)\) on the segments $s$ for three different types of curves -- namely, lines, circles/arcs, and all other types of curves grouped in \textit{general} category. First, a local PCA frame \cite{Shlens2014ATOPCA} is derived on the curve segments \(s\) using their arc-length parameterization. Next, the magnitude of displacement fields is dependent on a single input skill parameter $\kappa$ defining five different levels (L1 -- L5) of sketch artist -- \textit{e.g.,} hand drawing level to professional artist.  (see \cref{fig:A2Z_Sketching} and more results in the supplement).  
\vspace{-0.4cm}
\paragraph{Local planar frame.}
For each BRep edge \(e\in \boldsymbol{\mathcal{E}}\) with a set of vertices \(\{\boldsymbol{x}_i\}_{i=1}^{N}\), its local PCA frame
\(\{\boldsymbol{e}_0,\boldsymbol{e}_1,\boldsymbol{e}_2\}\) has the origin \(\boldsymbol{o}=\frac{1}{N}\sum_i \boldsymbol{x}_i\).
The axis \(\boldsymbol{e}_0\) aligns with the principal direction of the edge (tangent).
Any point on the edge is mapped to the local 2D coordinates ($
t=\boldsymbol{e}_0^\top(\boldsymbol{x}-\boldsymbol{o})$,\, $u=\boldsymbol{e}_1^\top(\boldsymbol{x}-\boldsymbol{o})$) that are 
distorted in this flat \((t,u)\) domain by
\(
\tilde{\boldsymbol{x}}=\boldsymbol{o}+(t+\Delta t)\,\boldsymbol{e}_0+(u+\Delta u)\boldsymbol{e}_1
\) (or by polar updates \(\Delta r,\, \Delta\theta\) for circular/arc type curves).


%
%
\begin{figure}[t]
  \centering
  \includegraphics[width=0.99\linewidth]{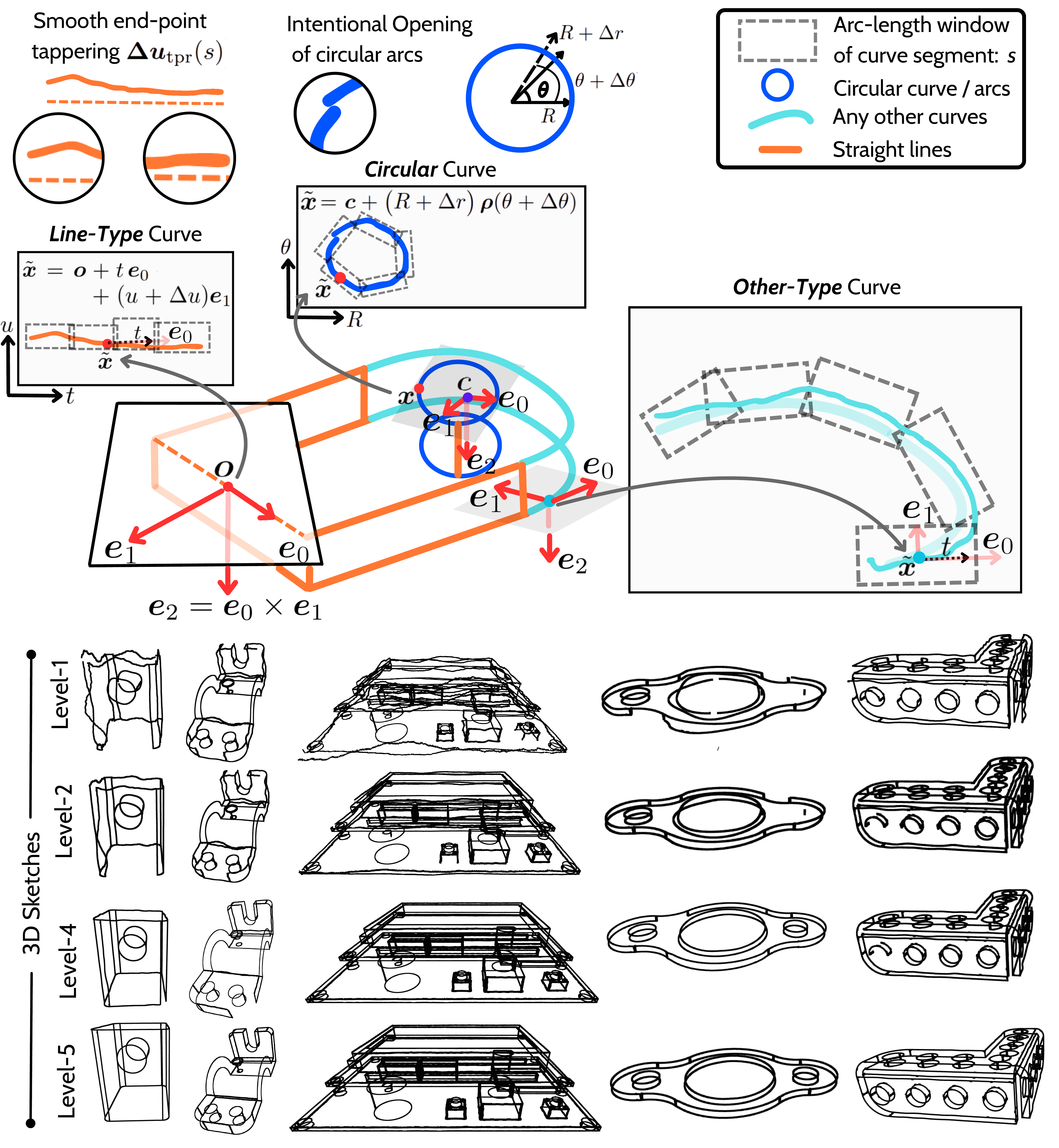}
   \vspace{-0.3cm}
   \caption{\textit{Five} million sketches with BRep annotations}
   \label{fig:A2Z_Sketching}
\end{figure}
\vspace{-0.4cm}
\paragraph{Skill parameter of a sketch artist.}
A single scalar \(\kappa\in\{1,\dots,5\}\) modulates \enquote{hand skill}: lower \(\kappa\) \(\Rightarrow\) higher displacement amplitudes, more arc-length segments of a curve of length $L$, and more jitter. We use
\vspace{-0.2cm}
\[
\alpha(\kappa)=\frac{1}{5}\left(6-\kappa\right)\in[0,1],
\vspace{-0.1cm}
\]
and scale deviations by a base magnitude \(A_0(\kappa,L)=\alpha(\kappa)\,c_L\,L\) with \(c_L\!\in\![10^{-3},10^{-2}]\) as a function parameter for different displacement fields according to skill.
%
%

\vspace{0.1cm}
\noindent\textbf{Straight line segments.}
For line type curves, the displacement field \(\boldsymbol{\Delta \Vec{u}}_{\text{line}}\) contains a mean reverting random walk \cite{MeanReverting_BM} \(\boldsymbol{\Delta u}_{\text{mr}}\), end segment tapering \(\boldsymbol{\Delta u}_{\text{tpr}}\), and a frequently bow-shaped bulge \cite{Windowbow_Sturgis, WindowBow78} \(\boldsymbol{\Delta u}_{\text{bow}}\) fields along the \(u\)-direction and is capped within \(C\in\left[-c,c\right]\). Next, a light tangential jitter \(\boldsymbol{\Delta t}_{\text{jitter}}\) is applied along the \(t\) direction. Finally, all displacement fields are smoothed using a moving average \(S_k\), resulting in \(\boldsymbol{\Delta \Vec{u}}_{\text{line}}\):  
\vspace{-0.2cm}
\begin{equation}
=
\begin{bmatrix}
S_k\!\left(\operatorname{clip}_c\!\big(
\boldsymbol{\Delta u}_{\mathrm{mr}}(s)+
\boldsymbol{\Delta u}_{\mathrm{bow}}(s)\big)
+\boldsymbol{\Delta u}_{\mathrm{tpr}}(s) \right)\\[4pt]
\boldsymbol{\Delta t}_{\mathrm{jitter}}(s)
\end{bmatrix}
\label{eq:LineSegment_Jitter}
\end{equation}
\vspace{-0.4cm}
\paragraph{Circular/arc segments.}
For circle/arc type curves, we first fit the best circle \(\left(\boldsymbol{c}, R, \theta\right)\) and re-parameterize it in polar form. Unlike line segments, several harmonics $H_\theta$ are sampled to create bumps. Some arc-length segments are skipped to apply intentional openings. 
We set \(\boldsymbol{\Delta} r=\boldsymbol{\Delta}\theta_{\text{jit}}=0\) at the start and end angles to avoid gaps. The radial and tangential jitters are linear projections onto tapered sinusoidal bases \(T(s)\) \cite{harris2005use}.
The points are lowered by
\(\tilde{\boldsymbol{x}}=\boldsymbol{c}+\big(R+\boldsymbol{\Delta} r\big)\,
\boldsymbol{\rho}(\theta+\boldsymbol{\Delta} \theta_{\text{jit}})\),
with \(\boldsymbol{\rho}\) the radial direction of the unit in the plane, resulting in \(\boldsymbol{\Delta \Vec{u}}_{\text{arc}}(s) = \left(\boldsymbol{\Delta} r, \boldsymbol{\Delta} \theta_{\text{jit}}\right)^\texttt{T}\)
\vspace{-0.2cm}
\begin{equation}
=
\begin{bmatrix} \mathbf{B}(\theta) & \mathbf{0} \\ \mathbf{0} & \mathbf{V}(\theta) \end{bmatrix}
\begin{bmatrix} \mathbf{c}(s) \\ \mathbf{d}(s) \end{bmatrix} \text{where},
\end{equation}
\vspace{-0.3cm}
\noindent\[
\mathbf{B}=
\begin{aligned}[t]
&\bigl[\cos(\theta+\phi_{1}),\ \tfrac{1}{2}\sin(\theta+\phi_{2}),\\
&\quad \{\sin(k\theta+\psi_k),\ 0.6\cos(k\theta+0.73\psi_k)\}_{k=2}^{H_\theta}\bigr],
\end{aligned}
\]
$\mathbf{c}(s)=\!\big[A\alpha_{1},\,A\alpha_{2},\,\{a_kT(s),\,a_kT(s)\}_{k=2}^{H_\theta}\big],\,$
$\mathbf{V}=\!\{\sin(k\theta{+}\eta_k)\}_{k=1}^{H_\theta},\;$ and 
$\mathbf{d}(s)=\!\{\beta_kT(s)\}_{k=1}^{H_\theta}.
$
This vector form preserves roundness while adding low-frequency wobble and tapered higher harmonics, as described in \cite{harris2005use}. The curve bulging amplitudes $\alpha_1,\alpha_2\sim U\left[0.6, 1.0\right]$. Similarly, the phase shifts $\phi_1, \phi_2$ for lower harmonics and $\eta_k, \psi_k$ for the higher harmonics adjust orientation of the sketching bulge. 

\vspace{-0.4cm}
\paragraph{Other General curves.}
For ellipses, B-Splines, or unknown curve types, we define a local arc-length segment and $K$ number of small windows to apply multiple bows \cite{Windowbow_Sturgis, WindowBow78}, mean reversion, and tapering  along the $u$-direction, as defined in \cref{eq:LineSegment_Jitter}. The overlapping displacement fields are averaged to estimate $\boldsymbol{\Delta \Vec{u}}_{\text{gen}}(s)$.  
\vspace{-0.1cm}
\begin{figure}[t]
  \centering
  \includegraphics[width=0.99\linewidth]{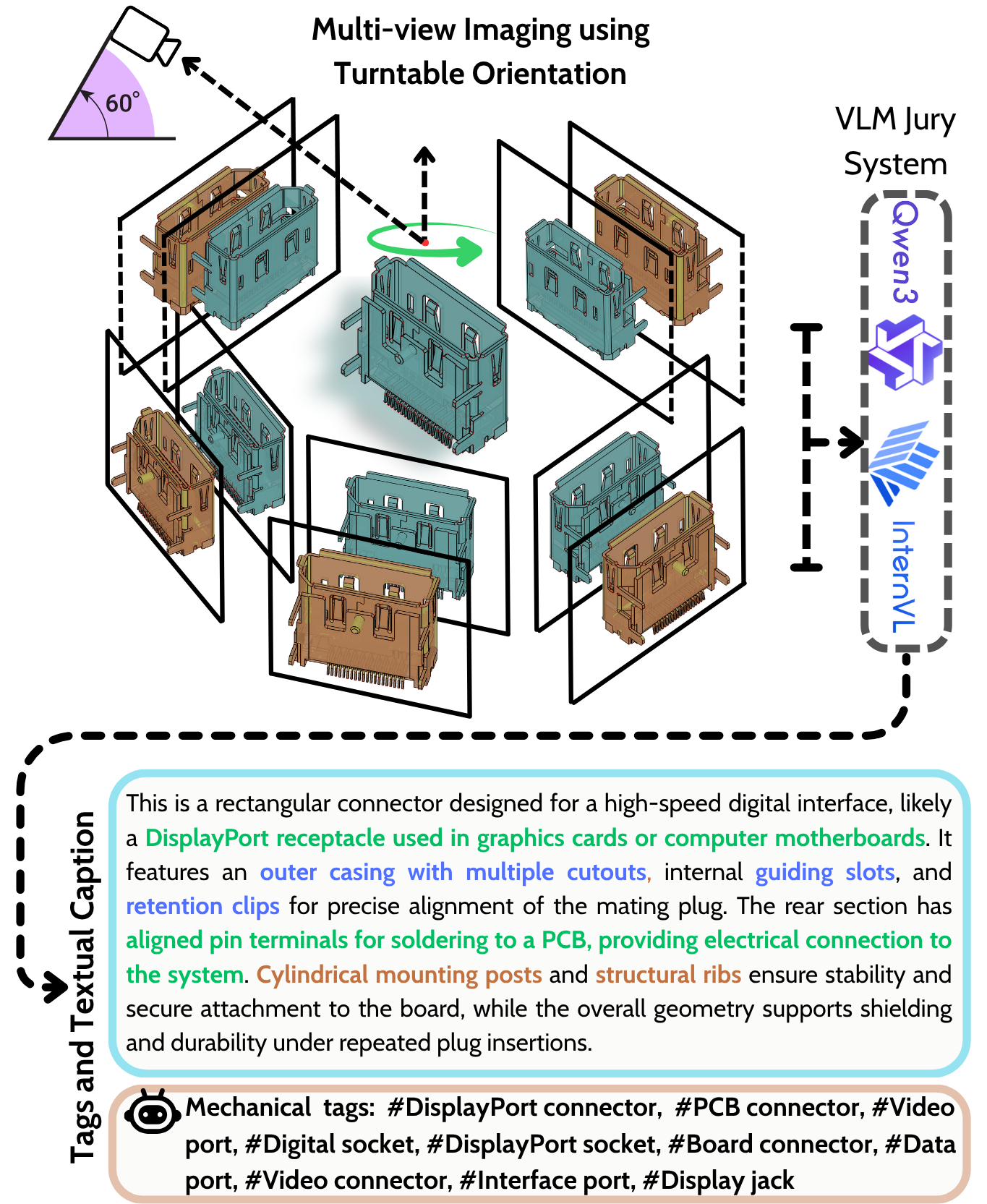}
  \caption{$\sim$5K models of charging points, serial bus etc.}
   \label{fig:MultiviewTurntable}
\end{figure}
\subsection{Multi-view Images, Captions and Tags}\label{subsec:3_AnnotTextImages}
High-fidelity 3D content generation from natural text prompts, \textit{i.e.,} text-to-3D \cite{MARVEL40M_2025_CVPR, ICML2024_Kabra, Clay2024, Cap3D2023}, has rapidly advanced gaming, fashion, AR/VR, and filming industries \cite{li2023generative}. However, apart from HoLA \cite{HoLABRep_2025}, no large-scale dataset provides visual elaborations of CAD models. Text2CAD \cite{khan2024text2cad}, CAD-MLLM \cite{xu2024cadMLLM}, and CADmium offer textual annotations only for \textit{sequential design history} models in DeepCAD \cite{DeepCAD_2021_ICCV}. In A2Z, we employ a novel jury system of two state-of-the-art VLMs, Qwen3-14B \cite{yang2025qwen3,chen2024farInternVL} and InternVL-26B \cite{chen2024internvl}, to generate high-quality captions and tags for all CAD models.  

Our caption pipeline uses \textit{twelve} multi-view images of each BRep, arranged in a 4x3 grid. As illustrated in \cref{fig:MultiviewTurntable}, a turntable system captures transparently rendered BRep views with uniform and face-type color coding. This rendering comprehensively exposes inner primitives (e.g., loops, holes, shells, boundaries), enabling the VLM jury to \textbf{\textit{(i)}} infer primitive details, \textbf{\textit{(ii)}} identify real-world resemblance and use, \textbf{\textit{(iii)}} highlight key surfaces, and \textbf{\textit{(iv)}} describe the overall 3D structure.  

The system also produces twenty concise tags per CAD model ($\leq$3 words each), serving as cross-references to industry terms. For example, tags like \textit{bolt cutout, fitting cutout, prismatic cutout, and thread count or cutout} allow efficient stocktaking and targeted asset management. Inspired by ImageNet \cite{ImageNet} and WordNet \cite{fellbaum1998wordnet}, we organize tags into six feature classes -- \textbf{(i)} Geometric Features, \textbf{(ii)} Miscellaneous, \textbf{(iii)} Mechanical Components, \textbf{(iv)} Structural Elements, \textbf{(v)} Electronic Components, and \textbf{(vi)} Fasteners -- followed by four successive tree-structured category layers for systematic traversal and mining of the A2Z database (see supplement for analytics).
\vspace{-0.1cm}
\subsection{Annotating Electronic Enclosures}\label{sec:3_ElectronicEnclosureData}
\noindent Three skilled designers created \(\sim\)20K CAD models in Onshape \cite{onshape} at \(\sim\)900 designs day, focusing on corner-rich parts of tablet and smartphone type casings. 
To design such models, an algorithmic variation of spline control points was created around the BRep faces at the corners, generating diverse geometries by preserving endpoint coincidence and \(G^{1}/G^{2}\) continuity constraints. Additionally, \(\sim\)5K charging ports, sockets, and casings were also modeled to expand A2Z and strengthen CAD reverse engineering for high-demand categories. More results are also in the supplement. 

\vspace{-0.1cm}
\section{Foundation Models for BRep Learning}
\label{sec:4_GeometricDL_Models}
\vspace{-0.15cm}
\paragraph{Boundary and Junction Detection.}\label{subsec:4_BJDetection} We have implemented a baseline neural model with DGCNN \cite{dgcnn} and two heads for detecting boundary and corner vertices, similar to ComplexGen \cite{guo2022complexgen}, PieNet \cite{wang2020pie}, and BRepDetNet \cite{BRepDetNet_2024ICMI}, as a classical binary classification task. This is considered the most important step in training for other BRep primitive parsing and parametric fitting tasks \cite{yan2021hpnet,spfn,le2021cpfn,sharma2020parsenet,BRepSeg_2025_TIP}. We randomly selected 300K CAD models from the first and last thirty chunks of ABC and used our A2Z annotations to train the geometric deep learning model on two Nvidia H100 GPUs for just 20 epochs over four days. Our network architecture is similar to that of \cite{BRepDetNet_2024ICMI} and \cite{wang2020pie}, with DGCNN \cite{dgcnn} serving as the backbone module for point cloud encoding. Subsequently, the per-point binary classification task between boundary and non-boundary members, as well as the focal loss function, is also similar to those suggested in \cite{wang2020pie, BRepDetNet_2024ICMI} (for more details, see the supplement).
\vspace{-0.2cm}
\section{Benchmarking and Evaluation}
\label{sec:5_Experiments_Evaluation}
\vspace{-0.1cm}
We have carried out a thorough evaluation process to quantify the quality factor of our \textbf{A2Z} dataset and its annotations. Subsequently, we benchmark the performance of our foundation model against the baselines.
%
%
\subsection{Quantitative Assessment on Dataset}\label{subsec:DataEval}
\noindent\textbf{3D Scan and Sketch Annotation Assessment.} We assessed the visual quality of 3D scan annotations across BRep face IDs, face types, and 3D sketches of varying skill levels using three evaluators -- \textbf{\textit{(i)}} human feedback, \textbf{\textit{(ii)}} Gemini \cite{gemini}, and \textbf{\textit{(iii)}} GPT-5 \cite{achiam2023gpt}. From 100 chunks of A2Z, we randomly sampled \textbf{\textit{10K}} CAD models for this evaluation. A parallel visualization tool was developed to explore annotated 3D scans and original BRep files, and ten human participants rated the quality from 1 (poor) to 10 (best). Evaluation sessions were recorded for cross-reference, and the saved images were input into Gemini and GPT-5 for rating on the same scale. As shown in \cref{tab:visual_quality_scores}, GPT-5 scored face IDs and face types slightly higher than humans and Gemini. However, the overall consensus was consistent. For 3D sketches, Gemini and GPT scores rose only marginally between artist levels 2–5, while humans showed a clearer appreciation of artistic nuances (\textit{More in the supplement}).
\begin{table}[t]
  \centering
  \small
  \setlength{\tabcolsep}{3.5pt}
  \renewcommand{\arraystretch}{0.9}
  \begin{tabular}{@{}l l ccc@{}}
    \toprule
    Method & \shortstack[l]{Sketch Artist\\Level} & Gemini & GPT & Human \\
    \midrule
    Face IDs  & \textemdash{} & 8.43 & 8.79 & 8.37 \\
    Face Type & \textemdash{} & 7.88 & 8.71 & 8.05 \\
    \multirow{4}{*}{Sketches} & 2 & 8.21 & 7.47 & 8.36 \\
                              & 3 & 7.87 & 7.71 & 9.12 \\
                              & 4 & 7.97 & 8.11 & 9.03 \\
                              & 5 & 8.31 & 8.43 & 9.61 \\
    \bottomrule
  \end{tabular}
  \vspace{-0.3cm}
  \caption{Visual quality scores (higher is better).}
  \label{tab:visual_quality_scores}
\end{table}

\begin{table}[t]
  \centering
  \scriptsize
  \setlength{\tabcolsep}{2.5pt}
  \renewcommand{\arraystretch}{0.9}

  \begin{tabular}{c c c}
    \begin{minipage}{0.5\columnwidth}
      \centering
      \begin{tabular}{@{}lccc@{}}
        \toprule
        \textbf{Level} & \textbf{MLTD} & \textbf{Unigram} & \textbf{Bigram} \\
        \midrule
        Low  & 59.32 &  856 & 11355 \\
        High & 70.52 &  908 & 12181 \\
        \bottomrule
      \end{tabular}
      \vspace{2pt}

      {\scriptsize \textbf{(a) Avg. MLTD \cite{MARVEL40M_2025_CVPR} and uni/bigrams.}}
    \end{minipage}
    &
    \hspace{0.1em}\vrule\hspace{-0.1em}
    &
    \begin{minipage}{0.5\columnwidth}
      \centering
      \begin{tabular}{@{}lccc@{}}
        \toprule
        Method & ID & Type & Loop \\
        \midrule
        Boundary & 99.37 & 97.67 & 99.99 \\
        Face     & 99.93 & 99.83 & \textemdash{} \\
        \bottomrule
      \end{tabular}
      \vspace{2pt}

      {\scriptsize \textbf{(b) Annotation coverage} (\%).}
    \end{minipage}
  \end{tabular}
    \vspace{-0.3cm}
  \caption{Two tables highlights the lexical diversity (\textit{left}) on captions and 3D scans' annotation coverage statistics (\textit{right}).}
  \label{tab:mtld_and_coverage_side_by_side}
\end{table}
\vspace{0.1cm}
\noindent\textbf{Statistics on Annotation Coverage.} We quantify the completeness of 3D annotations as the percentage of accurate IDs of boundaries and faces present in our scan annotations (i.e., per vertex) compared to the ground truth CAD models (\textit{i.e.,} topological walk using \lq.step\rq~files). This coverage is measured by averaging over the entire dataset. \cref{tab:mtld_and_coverage_side_by_side}-(b) shows that we have almost perfect boundary coverage (ID: 99.37\%, Type: 97.67\%, Loop: 99.99\%) as well as face coverage (ID: 99.93\% and Type: 99.83\%).

\vspace{0.1cm} 
\noindent\textbf{Textual Caption Annotation Assessment.} Now, to assess the quality of our large corpus of textual annotations, we use the same 10K CAD samples and their captions (including tags). Next,   \cref{tab:mtld_and_coverage_side_by_side}-(a), we report n-gram statistics and lexical diversity (MLTD), following the protocol adopted in MARVEL-40M+\cite{MARVEL40M_2025_CVPR}. To make a clear distinction between captions generated by a single small vision language model (SLM)  Qwen-7B\cite{yang2025qwen3} and our jury system of two large VLMs, InternVL3-38B \cite{chen2024internvl} and Qwen3-30B, we evaluate the metrics on both sets (see \cref{tab:mtld_and_coverage_side_by_side}-(a)). 
\begin{figure*}[!t]
  \centering
\includegraphics[width=\linewidth]{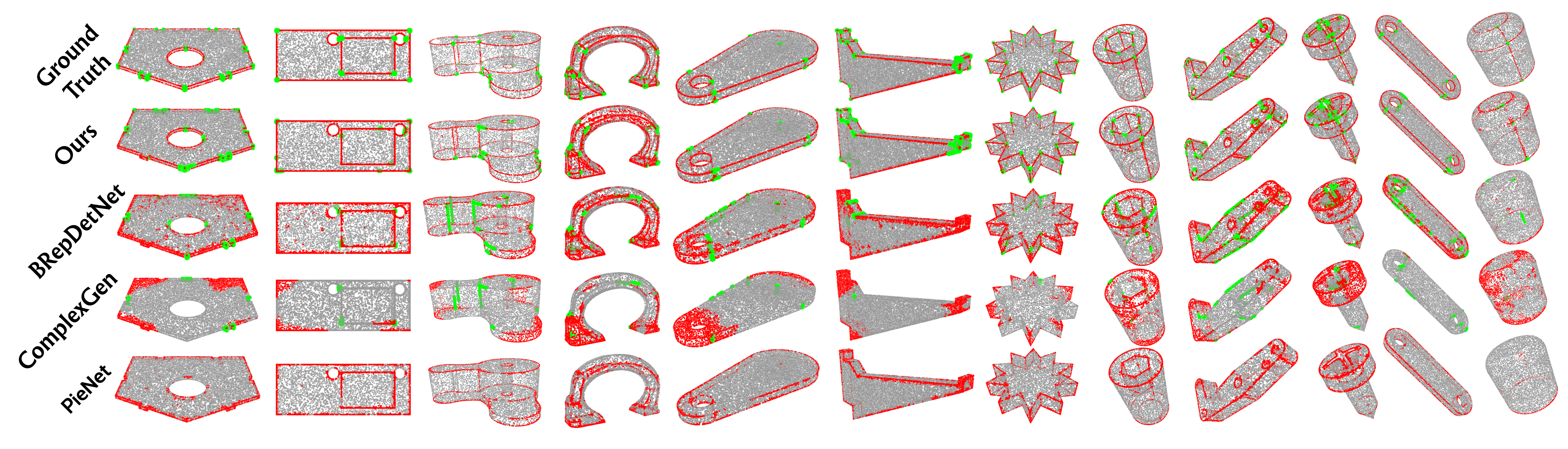}
    \vspace{-0.8cm}
   \caption{\textbf{Qualitative Comparison} of state-of-the-art methods in BRep \textit{Boundary} and \textit{Junction} classification as a downstream task.}
   \label{fig:Quality_BJ_Visual}
\end{figure*}
We also observe that high-quality annotations (using InternVL3-38B\cite{chen2024internvl} and Qwen3-30B\cite{yang2025qwen3}) yield a significantly better MLTD score (mean 70.52) compared to the low-quality condition (mean 59.33), resulting in a relative gain of 18.9\%. While unigrams show a minor increase, there is a more significant increase in bigrams in high-quality annotations compared to low-quality ones. High-quality annotations clearly represent a more diverse corpus, as captured by the statistics. 
\subsection{Benchmarking Foundation Model}\label{subsec:ModelEval}
\begin{table}[!t]
    \centering
    \small
    \setlength{\tabcolsep}{2pt}
    \renewcommand{\arraystretch}{1.0}
    \begin{adjustbox}{max width=\linewidth}
    \begin{tabular}{l *{8}{c}}
        \toprule
        \multirow{2}{*}{\textbf{Model}} &
        \multicolumn{4}{c}{\textbf{A2Z Chunks Seen (0--30, 70--99)}} &
        \multicolumn{4}{c}{\textbf{A2Z Chunks Unseen (31--69)}} \\
        \cmidrule(lr){2-5}\cmidrule(lr){6-9}
        & \multicolumn{2}{c}{\textbf{Boundary}}
        & \multicolumn{2}{c}{\textbf{Junction}}
        & \multicolumn{2}{c}{\textbf{Boundary}}
        & \multicolumn{2}{c}{\textbf{Junction}} \\
        \cmidrule(lr){2-3}\cmidrule(lr){4-5}\cmidrule(lr){6-7}\cmidrule(lr){8-9}
        & \textbf{Recall} & \textbf{Precision}
        & \textbf{Recall} & \textbf{Precision}
        & \textbf{Recall} & \textbf{Precision}
        & \textbf{Recall} & \textbf{Precision} \\
        \midrule
        \textbf{OURS}
        & \best{0.978} & \best{0.901}
        & \best{0.732} & \best{0.891}
        & \best{0.971} & \best{0.892}
        & \best{0.657} & \best{0.769} \\
        \textbf{BRepDetNet*}
        & \secondbest{0.903} & 0.781
        & \secondbest{0.454} & 0.561
        & \secondbest{0.873} & 0.701
        & \secondbest{0.353} & 0.426 \\
        BRepDetNet
        & 0.593 & 0.511
        & 0.355 & 0.231
        & 0.571 & 0.529
        & 0.311 & 0.226 \\
        \textbf{ComplexGen*}
        & 0.551 & 0.750
        & 0.297 & \secondbest{0.592}
        & 0.521 & 0.646
        & 0.212 & \secondbest{0.465} \\
        ComplexGen
        & 0.251 & 0.450
        & 0.097 & 0.288
        & 0.233 & 0.417
        & 0.055 & 0.310 \\
        \textbf{PieNet*}
        & 0.832 & \secondbest{0.885}
        & \na & \na
        & 0.782 & \secondbest{0.813}
        & \na & \na \\
        PieNet
        & 0.738 & 0.840
        & \na & \na
        & 0.699 & 0.806
        & \na & \na \\
        \bottomrule
    \end{tabular}
    \end{adjustbox}
    \vspace{-0.3cm}
    \caption{\textbf{Recall and Precision on A2Z}: Chunks Seen (0--30, 70--99) vs. Unseen (31--69). 
    \colorsquare{Best} = best,
\colorsquare{Second} = second best,
\colorsquare{NAcol} = N/A.
The asterisk \textbf{*} denotes freshly trained models on \textbf{A2Z}.}
    \label{tab:a2z_seen_unseen}
\end{table}
%
%
%
\begin{table}[!t]
    \centering
    \small
    \setlength{\tabcolsep}{2pt}
    \renewcommand{\arraystretch}{1.0}
    \begin{adjustbox}{max width=\linewidth}
    \begin{tabular}{l *{8}{c}}
        \toprule
        \multirow{2}{*}{\textbf{Model}} &
        \multicolumn{4}{c}{\textbf{CC3D Batches (0--10), 35K samples}} &
        \multicolumn{4}{c}{\textbf{Electronic Enclosure Chunk--100}} \\
        \cmidrule(lr){2-5}\cmidrule(lr){6-9}
        & \multicolumn{2}{c}{\textbf{Boundary}}
        & \multicolumn{2}{c}{\textbf{Junction}}
        & \multicolumn{2}{c}{\textbf{Boundary}}
        & \multicolumn{2}{c}{\textbf{Junction}} \\
        \cmidrule(lr){2-3}\cmidrule(lr){4-5}\cmidrule(lr){6-7}\cmidrule(lr){8-9}
        & \textbf{Recall} & \textbf{Precision}
        & \textbf{Recall} & \textbf{Precision}
        & \textbf{Recall} & \textbf{Precision}
        & \textbf{Recall} & \textbf{Precision} \\
        \midrule
        \textbf{OURS}
        & \best{0.961} & \best{0.854}
        & \best{0.633} & \best{0.810}
        & \secondbest{0.934} & \best{0.898}
        & \best{0.725} & \best{0.873} \\
        \textbf{BRepDetNet*}
        & \secondbest{0.763} & 0.807
        & 0.137 & 0.417
        & \best{0.950} & 0.856
        & \secondbest{0.410} & 0.787 \\
        BRepDetNet
        & 0.522 & 0.418
        & \secondbest{0.410} & 0.392
        & 0.610 & 0.602
        & 0.381 & 0.388 \\
        \textbf{ComplexGen*}
        & 0.427 & 0.743
        & 0.062 & \secondbest{0.437}
        & 0.774 & 0.809
        & 0.165 & \secondbest{0.815} \\
        ComplexGen
        & 0.280 & 0.444
        & 0.011 & 0.388
        & 0.221 & 0.491
        & 0.028 & 0.320 \\
        \textbf{PieNet*}
        & 0.663 & \secondbest{0.843}
        & \na & \na
        & 0.680 & \secondbest{0.870}
        & \na & \na \\
        PieNet
        & 0.481 & 0.713
        & \na & \na
        & 0.610 & 0.410
        & \na & \na \\
        \bottomrule
    \end{tabular}
    \end{adjustbox}
    \vspace{-0.3cm}
    \caption{Recall and Precision on \textbf{CC3D}~\cite{cc3d} and \textbf{Electronic Enclosure} in the \textbf{zero-shot} setting.}
\label{tab:cc3d_tablets_seen_unseen}
\end{table}
\vspace{-0.15cm}
To assess the fidelity of BRep-to-scan supervision, we evaluated the performance of our proposed foundation model (\cref{subsec:4_BJDetection}) against PIE‑Net \cite{wang2020pie}, ComplexGen \cite{guo2022complexgen}, and BRepDetNet \cite{BRepDetNet_2024ICMI} using their model check-points as well as a freshly trained model on our \textbf{A2Z} labels for boundary and corner detection tasks. CAD categories and types vary widely across 100 A2Z chunks, so performance is reported on the test sets ($\sim$2K samples per chunk) over “seen’’ (0–30, 70–99) and “unseen’’ (31–69) chunks using recall–precision for boundary and junction vertices (see \cref{tab:a2z_seen_unseen}). Training utilized uniformly sampled 10K scan points per shape and a batch size of 8 for the binary classification of boundary versus non-boundary. For the junction detection task, the process is conducted on top of the detected boundary points, following the same protocol. boundary vertices are sparse relative to all scan points, and junction vertices are even sparser relative to the boundaries, making corner detection particularly challenging for generic classifiers. Thanks to our high-quality annotations, the proposed model consistently achieves the highest recall–precision balance for both tasks on A2Z chunks (\cref{tab:a2z_seen_unseen}), on the CC3D dataset that was never used for training (\cref{tab:cc3d_tablets_seen_unseen}), and on the newly added chunk for electronic enclosures. 

When models from baseline methods are freshly trained, we see that \cite{guo2022complexgen} and \cite{BRepDetNet_2024ICMI} gain \textit{recall-precision} in the ranges of $\sim$(10\%, 30\%) on the boundary task and $\sim$ (4\%, 33\%) on the junction task. This is due to the remarkable quality of our annotations. The performance gain of PieNet \cite{wang2020pie} is no different from that of other methods on the boundary task. However, PieNet training and inference consistently fail in the junction task. Notably, the drop from seen to unseen chunks is smaller for our proposed approach than for the baselines. Qualitative results in \cref{fig:Quality_BJ_Visual} demonstrate cleaner boundary classification and more complete corner recovery by our model, with fewer spurious edges on planar faces and fewer missed junctions on \textbf{A2Z} test set. Collectively, the figure and table support the claim that the proposed foundation model can reliably transfer BRep topology cues to scans for downstream CAD reconstruction.

\vspace{-0.2cm}
\section{Conclusion and Future Work.}\label{sec:6_Conclusion_FW}
\vspace{-0.2cm}
This work introduces a unique, most complete, and largest dataset of CAD and scan pairs with multimodal annotations. We conducted a thorough benchmarking of ground-truth annotations using Human feedback, GPT-5-based scoring, and the most suitable metrics to quantify the quality. Our foundation model and state-of-the-art results in boundary and junction classification demonstrate the value of this dataset for developing numerous new applications in generative BRep modeling, reverse engineering, and mining CAD features.

\section{Acknowledgment}
This project was primarily funded by BITS Pilani Hyderabad’s NFSG Grant (Reference N4/24/1033). We are thankful to Vinci4D.ai for providing some CAD model samples of electronic enclosures. 
{
    \small
    \bibliographystyle{ieeenat_fullname}
    \bibliography{main}
}

\maketitlesupplementary

\makeatletter
\let\addcontentsline\SavedAddContentsLine
\makeatother

\setcounter{section}{0}
\renewcommand\thesection{\Alph{section}}
\renewcommand\thesubsection{\thesection.\arabic{subsection}}

\setcounter{tocdepth}{2}
\renewcommand{\contentsname}{\LARGE\bfseries Contents}



%
%
%
\tableofcontents
\bigskip

We structure the supplementary material according to the content outlined above. 
\vspace{-0.4cm}
\section{Multi-scale SPH for Neighbor Labeling}\label{sec:Supply_ProximAware_Annot} 
In A2Z, we have generated 3D meshes that closely resemble real-world scanning outcomes. This process is carried out by uniquely combining multiple computational geometry algorithms, guided by BRep geometry and topology features.   Many uncommon problems arise when too many small faces are adjacent to one another. These corner cases create confusion about how to assign the nearest neighbor. Let \(\mathcal{P}=\{\boldsymbol{p}_i\}_{i=1}^{N}\) be the scan, and let \(\boldsymbol{x}\in\mathcal{S}_{E}\cup\mathcal{S}_{F}\) be a sampled BRep point with local frame \(R(\boldsymbol{x})=[\boldsymbol{t},\boldsymbol{u},\boldsymbol{n}]\) (tangent, in–plane, normal).

\subsection{Coedge Length–Aware Neighbors.}\label{supply:subsec:Co-edge-LengthAware_Neighbor}
\vspace{-0.15cm}
For a boundary sample \(\boldsymbol{x}\in\mathcal{S}_{E}\) with parent edge length \(L_{e(\boldsymbol{x})}\), we first define a base scale $
h_{E}(\boldsymbol{x})=\alpha_{E}\,L_{e(\boldsymbol{x})},\,
\mathcal{R}_{E}(\boldsymbol{x})=\{\,r_k(\boldsymbol{x})=\gamma_k h_{E}(\boldsymbol{x})\,\}_{k=1}^{K},$
and then for a BRep face sample point \(\boldsymbol{x}\in\mathcal{S}_{F}\) with grid pitch \(\Delta_f\), the same base scale for the face is defined as:
\[
h_{F}(\boldsymbol{x})=\alpha_{F}\,\Delta_f \, \text{and}\, 
\mathcal{R}_{F}(\boldsymbol{x})=\{\,r_k(\boldsymbol{x})=\gamma_k h_{F}(\boldsymbol{x})\,\}_{k=1}^{K}.
\]
Next, an anisotropic metric along the BRep edge measures whether small gaps exist of the two adjacent faces are kind of "seam edges",
\[
\begin{aligned}
d_i(\boldsymbol{x}) &= 
\big\|\boldsymbol{p}_i-\boldsymbol{x}\big\|_{A(\boldsymbol{x})}
= \sqrt{\big(\boldsymbol{p}_i-\boldsymbol{x}\big)^{\!\top} A(\boldsymbol{x})\big(\boldsymbol{p}_i-\boldsymbol{x}\big)},\\
A(\boldsymbol{x}) &= 
R(\boldsymbol{x})\,
\operatorname{diag}\!\big(\sigma_t^{-2},\sigma_u^{-2},\sigma_n^{-2}\big)\,
R(\boldsymbol{x})^{\!\top}.
\end{aligned}
\]
with \(\sigma_t\!>\!\sigma_u\!\ge\!\sigma_n\) on the edges and \(A(\boldsymbol{x})\propto I\) on the BRep faces. Once 

\vspace{-0.2cm}
\subsection{SPH kernel with BRep Surface features.}\label{supply:subsec:BRepSurfaceFeat_Neighbor}
We use the cubic–spline SPH kernel in 3D,
\[
W(q,h)=\frac{1}{\pi h^{3}}
\begin{cases}
1-\tfrac{3}{2}q^{2}+\tfrac{3}{4}q^{3}, & 0\le q<1,\\[2pt]
\tfrac{1}{4}(2-q)^{3}, & 1\le q<2,\\[2pt]
0, & q\ge 2,
\end{cases}
\]
with \(q=d_i(\boldsymbol{x})/h\).
Non–sharp operational features (filets, chamfers, bosses) receive a curvature–aware smoothing length
\[
\begin{aligned}
h(\boldsymbol{x}) &= \eta\,\min\!\Big(h_{*}(\boldsymbol{x}),\,R_{\kappa}(\boldsymbol{x})\Big),\\
R_{\kappa}(\boldsymbol{x}) &= \frac{1}{\max\big(|\kappa_1(\boldsymbol{x})|,\,|\kappa_2(\boldsymbol{x})|,\,\varepsilon\big)}.
\end{aligned}
\]
where \(h_{*}\!\in\!\{h_{E},h_{F}\}\) and \(\kappa_{1,2}\) are the principal curvatures from the B–Rep surface at \(\boldsymbol{x}\).

Optionally, a normal–consistency gate improves robustness near joints:
\[
g_i(\boldsymbol{x})=\exp\!\big(-\lambda\,[1-|\boldsymbol{n}(\boldsymbol{x})^{\!\top}\boldsymbol{n}_i|]\big),
\]
where \(\boldsymbol{n}_i\) is the scan normal at \(\boldsymbol{p}_i\) (if available). 
The anisotropic metric \(A(\boldsymbol{x})\) and the SPH kernel $W(q,h)$ jointly ensure coverage of close–proximity vertices and accurate recovery on smooth CAD features without overextending across sharp boundaries, as described in \textbf{Eq. 2} of the main matter. 
\section{Synthesized Sketch and Realism Analysis}
\label{sec:sketch_realism}
A critical modality in A2Z-10M+ is the \emph{sketch}, which bridges the gap between early-stage conceptual design and precise BRep geometry.
Sketches in engineering practice span a wide stylistic spectrum—from quick freehand concept doodles to dimensioned orthographic projections—and a dataset that targets sketch-based CAD retrieval and reconstruction must faithfully cover this range.
\begin{figure*}[!t]
  \centering
\includegraphics[width=0.99\linewidth]{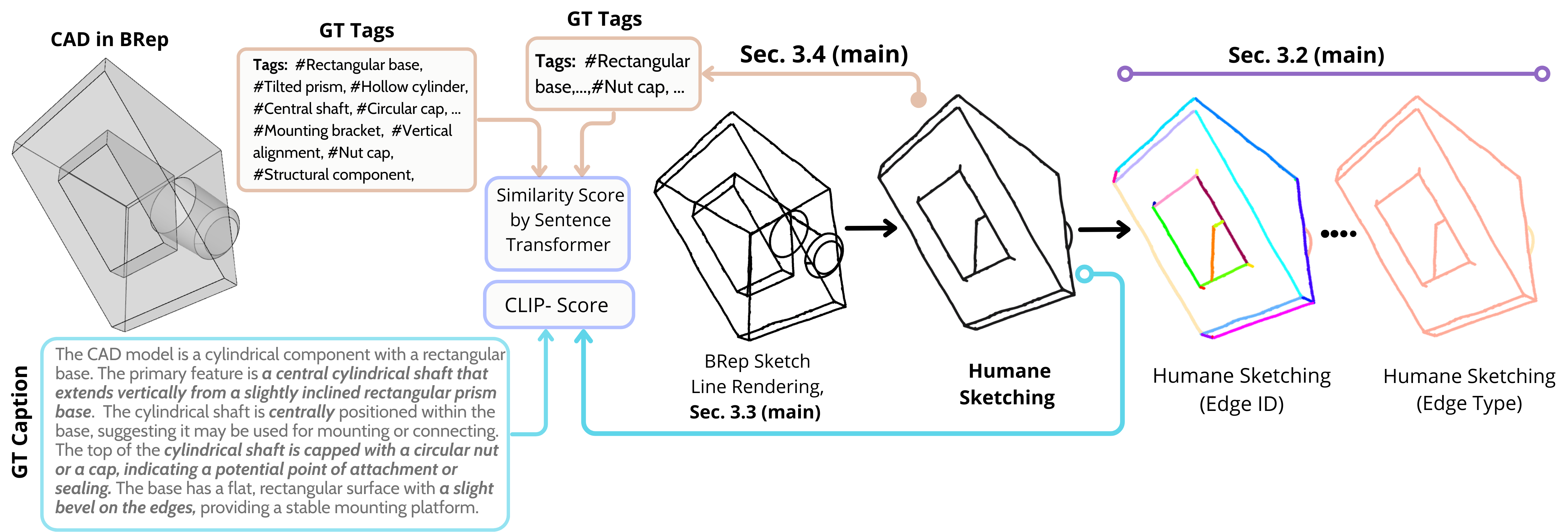}
   \vspace{-0.2cm}
   \caption{\textbf{Realism of Sketch.}: We compute a CLIP cosine-similarity score of \textbf{30.58} between generated captions and the corresponding 2D sketches, confirming that textual descriptions are well-grounded in visual content. Additionally, a tag-level similarity score of \textbf{64.7\%} (CAD-derived tags vs.\ sketch-derived tags) demonstrates that the hierarchical tag vocabulary remains consistent across the BRep and sketch modalities.
}
   \label{fig:SketchRealism}
\end{figure*}
\subsection{Generation Pipeline}
\label{subsec:sketch_pipeline}
A2Z adopts a multi-level 3D sketching framework (Sec.~3.3 of the main paper) that builds on recent advances in differentiable stroke rendering. Beginning from the native BRep representation, the pipeline proceeds through four stages:
\begin{enumerate}[leftmargin=*,itemsep=2pt,topsep=2pt]
    \item \textbf{View capture.} A set of canonical viewpoints is sampled around the CAD model, and perspective/orthographic projections are rendered, preserving the association between projected curves and their parent BRep edges.
    \item \textbf{Occlusion removal.} Hidden lines are identified via depth-buffer queries and either removed (for engineering-style drawings) or rendered with dashed strokes (for see-through conceptual sketches).
    \item \textbf{Loop pruning.} Redundant or visually cluttered silhouette loops are simplified while retaining topologically significant contours (e.g., holes, fillets, chamfer boundaries).
    \item \textbf{Stroke rendering.} The surviving curves are rasterized with stylistic variation (line weight jitter, over-sketching, slight positional noise) to approximate hand-drawn appearance at multiple fidelity levels.
\end{enumerate}
The result is a hierarchy of sketch abstractions—referred to as \emph{Level-2} through \emph{Level-5} in decreasing order of realism—each retaining the semantic edge-type and edge-ID labels inherited from the parent BRep.
This label preservation is essential: while pixelated sketches are sufficient for image-based \emph{retrieval}, they cannot drive BRep \emph{reconstruction} without the accompanying semantic edge annotations that specify curve type (line, arc, B-spline), connectivity, and face adjacency.

\subsection{Quantitative Realism Evaluation}
\label{subsec:sketch_realism_eval}

To validate sketch realism, we conducted a perceptual evaluation on a held-out set of 5\,000 newly generated 2D sketches (half of our evaluation pool, corresponding to ll.\,455--464 of the main paper).
Each sketch was scored on a 1--10 realism scale by three independent evaluators: two frontier vision-language models (GPT-5, Gemini) acting as automated raters, and a panel of mechanical engineering graduate students serving as human raters.
Table~\textcolor{cvprblue}{6} reports the mean and standard deviation per abstraction level.

\begin{table*}[t]
    \label{tab:sketch_real}
    \centering
    \Large
    \setlength{\tabcolsep}{5pt}
    \begin{tabular}{@{}llcccc@{}}
        \toprule
        \textbf{Evaluator} & \textbf{Modality} & \textbf{Level-2} & \textbf{Level-3} & \textbf{Level-4} & \textbf{Level-5} \\
        \midrule
        GPT-5  & sketch & $7.65 \pm 0.34$ & $7.20 \pm 0.48$ & $5.53 \pm 0.38$ & $5.05 \pm 0.39$ \\
        Gemini & sketch & $\mathbf{9.01 \pm 0.28}$ & $7.33 \pm 0.52$ & $6.06 \pm 0.58$ & $3.92 \pm 0.47$ \\
        Human  & sketch & $7.20 \pm 0.32$ & $\mathbf{7.86 \pm 0.38}$ & $7.85 \pm 0.44$ & $7.67 \pm 0.71$ \\
        \bottomrule
    \end{tabular}
    \caption{Realism scores (mean $\pm$ std, 1--10 scale) for \textbf{\textit{multi-level 2D sketches}} evaluated by automated VLM raters and human engineering students.  Higher is better.  Level-3 achieves the strongest evaluator consensus, balancing abstraction with recognizability.}
\end{table*}

\begin{table*}[t]
    \centering
    \label{tab:scan_realism}
    \vspace{-0.2cm}
    \Large
    \setlength{\tabcolsep}{4pt}
    \begin{tabular}{@{}llcccc@{}}
        \toprule
        \textbf{Evaluator} & \textbf{Modality} & \textbf{Step-I} & \textbf{Step-II} & \textbf{Step-III} & \textbf{Step-IV} \\
        \midrule
        GPT-5  & scan & $6.51 \pm 0.84$ & $6.54 \pm 1.01$ & $7.58 \pm 0.58$ & $\mathbf{9.41 \pm 0.60}$ \\
        Gemini & scan & $7.73 \pm 1.48$ & $7.91 \pm 0.92$ & $8.54 \pm 0.78$ & $\mathbf{9.28 \pm 0.81}$ \\
        Human  & scan & $6.32 \pm 0.29$ & $6.58 \pm 0.80$ & $7.20 \pm 0.51$ & $\mathbf{9.83 \pm 0.22}$ \\
        \bottomrule
    \end{tabular}
    \caption{Realism scores (mean $\pm$ std, 1--10 scale) for \textbf{\textit{synthesized scans}} at each pipeline stage, evaluated against Artec3D ground-truth scans of CC3D models. Refer to Figure.~\ref{fig:CC3DScanVsSynthesis} for visuals. \textbf{Step~IV} achieves near-perfect human consensus.}
\end{table*}

Several trends emerge.
First, \emph{Level-2} sketches, which retain the most geometric detail, receive high scores from automated raters but are judged slightly lower by humans, likely because the density of strokes can appear mechanical rather than hand-drawn.
Second, \emph{Level-3} achieves the strongest overall evaluator consensus: it retains enough salient contours for reliable part identification while introducing sufficient stylistic abstraction to appear natural.
Third, at the more abstracted \emph{Level-4} and \emph{Level-5}, human raters remain relatively tolerant (scores above 7.6), whereas automated raters penalize the loss of geometric detail more sharply; this gap highlights a known limitation of VLM-based perceptual metrics, which tend to favour high-frequency detail.

These results confirm that the multi-level sketching pipeline produces perceptually convincing outputs across the realism spectrum, supporting both downstream retrieval tasks (where coarse sketches suffice) and reconstruction tasks (where Level-2/3 sketches with full semantic labels are preferred).

\section{Synthesized Scan and Realism Analysis}
\label{sec:scan_realism}

The second critical modality in A2Z-10M+ is the \emph{synthesized 3D scan}, which simulates the output of real structured-light or laser-based acquisition devices applied to CAD parts.
High-fidelity synthetic scans are essential for training and evaluating scan-to-BRep reconstruction methods~\cite{Point2Primitive2025,Point2CAD_Liu2024_CVPR,guo2022complexgen}, yet prior datasets typically add only simple Gaussian noise to uniformly sampled point clouds—a poor proxy for the spatially correlated artefacts produced by physical sensors.

\subsection{Artec3D-Calibrated Scan Synthesis}
\label{subsec:scan_pipeline}

A2Z addresses this gap through a four-stage scan synthesis algorithm (Sec.~3.1 of the main paper) whose parameters are calibrated against ground-truth (GT) scans produced by the Artec3D virtual scanner.
Artec3D is a proprietary tool that replicates the optical path, sensor calibration curves, and opto-photonic noise characteristics of its commercial structured-light hardware; scans produced by this tool are therefore treated as \emph{real ground truth} for evaluation purposes.
The four stages introduce progressively more realistic artefacts:

\begin{enumerate}[leftmargin=*,itemsep=2pt,topsep=2pt]
    \item \textbf{Step~I\;(Uniform sampling \& base noise).}  Points are sampled uniformly on the BRep surface and perturbed with isotropic Gaussian noise calibrated to the Artec3D sensor's baseline repeatability specification.
    \item \textbf{Step~II\;(View-dependent density).}  A virtual sensor viewpoint is introduced, and point density is modulated by the local surface-to-sensor angle, replicating the reduced sampling density observed on oblique and self-occluded regions.
    \item \textbf{Step~III\;(Structured artefacts).}  Spatially correlated noise patterns—edge ringing, specular dropout, and inter-reflection ghosts—are layered according to material BRDF models and local curvature, emulating systematic scanner errors absent from i.i.d.\ noise models.
    \item \textbf{Step~IV\;(Full sensor simulation).}  The complete Artec3D opto-photonic pipeline is applied, including lens distortion, structured-light fringe decoding residuals, and outlier generation at depth discontinuities, yielding scans that closely match real GT.
\end{enumerate}

\subsection{Stage-Wise Realism Evaluation}
\label{subsec:scan_realism_eval}

To quantify the realism gain at each stage, we synthesized scans from CC3D CAD models whose Artec3D GT scans are available. We evaluated each stage using the same three-evaluator protocol described in Sec .~\ref{subsec:sketch_realism_eval}, \textit{i.e.,} automated VLM raters and human experts. The Table~\textcolor{cvprblue}{7} reports the results employing the visual comparison tool shown in Fig.~\ref{fig:SupplyQualityEval_Proc}.

The results reveal a clear monotonic improvement in perceived realism across stages.
Steps~I and~II, which model only point distribution and view-dependent density, receive modest scores (6.3--7.9), indicating that density variation alone is insufficient to fool either human or automated raters.
The introduction of structured artefacts in Step~III yields a noticeable jump (roughly +1 point across all evaluators), confirming that spatially correlated sensor errors are a primary perceptual cue for scan authenticity.
Finally, Step~IV achieves near-ceiling scores—$9.83 \pm 0.22$ from human raters—demonstrating that the full opto-photonic simulation produces scans virtually indistinguishable from real Artec3D acquisitions.

The extremely low variance at Step~IV for human raters ($\sigma = 0.22$) is particularly noteworthy: it indicates strong inter-annotator agreement that the synthesized scans are realistic, providing a high-confidence quality floor for downstream learning tasks such as boundary detection, surface segmentation, and BRep fitting.

\section{Prompt for CAD Description and Tags}
\label{sec:cad_prompt}
\vspace{-0.2cm}

We have structured our prompt template for a mixture of VLMs -- Qwen3-30B \cite{yang2025qwen3} and InternVL3 \cite{chen2024farInternVL} (see \textit{Section 3.4} of the main matter)—and defined a consistent textual description and tag set. In the prompt template, the following segments are present:
\begin{promptbox}[Overall LLM's Role]
``You are a senior mechanical design engineer with deep expertise in component identification and sourcing. You are shown 12 images of a single CAD model from multiple views, arranged in a grid. Your task is to analyse the geometry and produce a technical description and classification tags suitable for a professional engineering database".
\end{promptbox}
After defining the role, we fix the rules for the description in in-context prompt learning in three different parts: \textbf{\textit{(1)}} Description, \textbf{\textit{(2)}} Filtering and Constraints, and \textbf{\textit{(3)}} Desired Output Structure.

\subsection{Description}\label{suppl:subsec:PromptDescription}
Write a single, concise paragraph (maximum 150 words) that follows these guidelines:
\begin{enumerate}[leftmargin=*,itemsep=1pt,parsep=0pt,topsep=0pt]
    \item \textbf{Primary identification.} Use specific, standard mechanical
    terminology only when the geometry makes the class obvious; if the type is
    unclear, describe the geometry directly and do \emph{not} speculate about its industry role.
    \item \textbf{Justification and geometry.} Justify your identification by
    describing the key geometric features that support it. If you cannot
    confidently name the part, instead of describing its dominant geometric forms
    (cylinders, prisms, \dots) and their spatial relationships using CAD terms
    (concentric, axial, radial, \dots).
    \item \textbf{Feature details.} Mention mounting features, holes, cut-outs,
    bosses, slots, or surface transitions (filets, chamfers, \dots) that appear
    critical to the part's function.
    \item \textbf{Synthesis of views.} Ensure that the description is a cohesive synthesis of all views, explicitly noting overall symmetries and important spatial relationships.
    \item \textbf{Potential application:} Briefly state a plausible industrial application or use case, and the likely engineering function of the component based on its geometry.
\end{enumerate}
\subsection{Applying Filtering and Constraints}\label{supply:subsec:FilteringConstraints}
\begin{promptbox}[Filtering and Constraints]
We have implemented some restrictions and ethical filtering constraints for Qwen3-30B \cite{yang2025qwen3} to generate the final response. For instance,

\begin{enumerate}[leftmargin=*,label=\arabic*.,itemsep=1pt,parsep=0pt,topsep=0pt]
  \item Do \textbf{not} use canned template phrases such as
  ``This appears to be a \dots{} shaft coupling'' or
  ``The component is a \dots{} cylinder'' unless the object is genuinely that
  specific type. Be original in your wording.
  \item The final tag list must contain 20 items in a flat structure. Do not
  include headings like ``Category/Class:'' in the output.
  \item Avoid generic, overused terms (e.g., \emph{flange}, \emph{bracket},
  \emph{housing}) unless the geometry explicitly and unambiguously matches those
  features, and you are more than $90\%$ confident.
\end{enumerate}
\end{promptbox}
\vspace{-0.3cm}
\subsection{Output Format}\label{supply:subsec:OutputFormat}
The response must follow this exact structure:
\vspace{-0.15cm}
\begin{promptbox}[Output Structure]
\begin{verbatim}
Description: {Your paragraph here}
Tags:
- Tag 1
- Tag 2
- ...
- Tag 10
\end{verbatim}
\vspace{-0.3cm}
\end{promptbox}

% ...
%
%
\begin{figure*}[!t]
  \centering
  \includegraphics[width=0.99\linewidth]{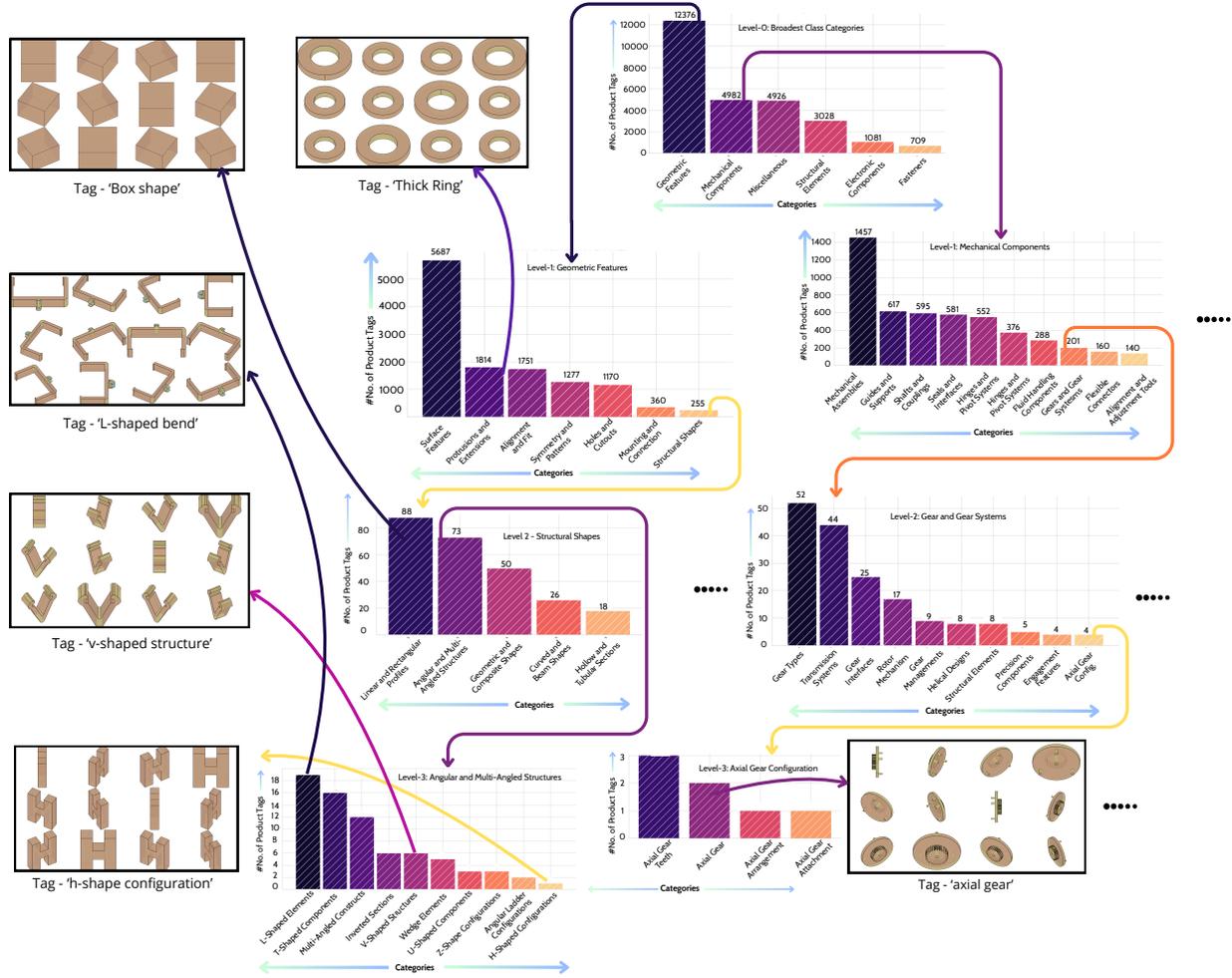}
   \caption{Hierarchical categorization of 1 million CAD models for retrieval and causal mining. The two examples illustrate how  -- \textbf{\textit{(on the right)}} one can fetch some CAD models when asked to find -- \lq A \textbf{\textit{mechanical component}} associated to \textbf{\textit{Gear type transmission}} systems supporting \textbf{\textit{Axial}} rotation\rq. }
   \label{fig:HierarchicalTag_Structure}
\end{figure*}

\section{Hierarchical Tags for CAD Retrieval}\label{supplysec:Hierarchical_Tags_Structure}
Geometric feature recognition \cite{BRepFormer25, REGASSAHUNDE2022100478} on BRep, and then streamlining those features to replace traditional part modeling techniques \cite{cherng1998feature}, can transform complex CAD retrieval for reuse purposes in design. Like many other important downstream tasks of BRep learning 
the retrieval and identification of CAD models by their geometric features \cite{BRepGAT_Lee_2023} automate manufacturing steps by saving time and reducing human error. For this purpose, a well-structured database of part components is necessary for ingestion. Such a database serves as a go-to reference that links the Computer-Aided Design (CAD) process to the Computer-Aided Manufacturing (CAM) workflow. Our methodology for this hierarchical mining and retrieval of CAD based on their captions and unique tags is described below:

\subsection{Semantic Hierarchy over 1M A2Z Models}
We begin with a corpus of roughly one million CAD models from the ABC dataset, each annotated with a visual caption and up to ten free-form tags, as mentioned in the main matter. An initial test run is first made on the $N=27{,}102$ distinct tags that appear throughout the corpus. We construct a semantic hierarchy that simultaneously reflects human design intuition and the empirical distribution of the data. We iteratively use Qwen3--30B \cite{yang2025qwen3} as an LLM to serve as a semantic classifier, running over $\lfloor\frac{N}{100}\rfloor$ unique tags in a chunk and then clustering them consistently across other chunks. Therefore, we have created semantic clustering with a maximum depth of 5 in distinct phases. 

\vspace{0.15cm}
\noindent\textbf{Phase-1: high-level assignment.}
Every tag $t$ is first routed into one of $K$ coarse parent categories $\{c_1,\dots,c_K\}$ (e.g., \emph{Geometric Features}, \emph{Mechanical Components}, \emph{Structural Elements}, \emph{Electronic Components}, \emph{Fasteners}, plus an \emph{Unclassified} bucket).  The model is prompted with the tag in isolation and returns a single category label, yielding a mapping
\vspace{-0.2cm}
\[
\phi_1 : t \longmapsto c_k.
\]
This step ensures that all later reasoning is constrained within a semantically coherent region of the space.

\vspace{0.1cm}
\noindent\textbf{Phase-2: recursive refinement.}
Within each coarse category, we recursively subdivide tag sets into more specific groups.  Given a set $S\subset\mathcal{T}$ of tags that share the same parent, the LLM is asked to propose between five and ten meaningful sub-groups that partition $S$ (e.g., under \emph{Geometric Features}, we obtain groups such as \emph{box-like shapes}, \emph{L-bends}, \emph{V-shaped structures}, and \emph{axial gears}, as shown in Fig.~\ref{fig:HierarchicalTag_Structure}).  The process is repeated for each child group until either (i) the group contains fewer than ten tags or (ii) a maximum depth is reached.  The result is an irregular, data-driven tree that is deep where the vocabulary is rich and shallow where it is sparse.

\vspace{0.1cm}
\noindent\textbf{Phase-3: attaching model identifiers.}
After the structure of the tree is fixed, we traverse it and for each leaf node $\nu$ we attach the set of CAD models
\vspace{-0.2cm}
\[
\mathcal{M}_\nu=\{\,m \in \mathcal{M} \mid \text{tag}(m)\ni t \text{ and } t\in \nu\,\}
\]
that carry any of the tags stored at $\nu$.  As a consequence, every model is reachable from multiple semantic leaves, coupling fine-grained vocabulary (e.g., ``\textit{axial gear with keyway}'') with the raw geometry of the associated CAD.

\vspace{0.15cm}
\noindent\textbf{Phase 4: statistical reporting.}
Finally, the fully assembled hierarchy is summarized to support downstream analysis and dataset curation.  Table~\ref{tab:root-distribution} reports the observed distribution of tags over the root categories, revealing, for instance, that nearly half of all tags are geometric in nature, while mechanical and structural terms jointly account for roughly one quarter of the vocabulary.  These statistics are later used to rebalance training sets and design retrieval strategies that avoid overfitting to overrepresented classes.
\begin{table}[t]
    \centering
    \caption{Distribution of the $27{,}102$ unique tags over top-level categories and their Percentage ($\%$) of CAD models.}
    \label{tab:root-distribution}
    \begin{tabular}{lrr}
    \toprule
    Category                & Tag Count & Percentage \\
    \midrule
    Geometric Features      & 12{,}230  & 45.13\% \\
    Unclassified            & 6{,}068   & 22.39\% \\
    Mechanical Components   & 4{,}601   & 16.98\% \\
    Structural Elements     & 2{,}724   & 10.05\% \\
    Electronic Components   & 801       & 2.96\%  \\
    Fasteners               & 678       & 2.50\%  \\
    \bottomrule
    \end{tabular}
\end{table}
\vspace{-0.2cm}
\subsection{Heterogeneous Graph Representation}\label{supply:subsec:HeterGraphMining}
\vspace{-0.15cm}
The hierarchical taxonomy is converted into a heterogeneous graph that unifies tags, categories, captions, and CAD models.  Let $\mathcal{M}$ denote the set of model nodes, $\mathcal{T}$ the set of tag nodes, and $\mathcal{C}$ the set of category nodes (internal and root).  The vertex set is
\[
V = \mathcal{M} \cup \mathcal{T} \cup \mathcal{C},
\]
and the edge set $E$ contains four main relation types:
\begin{enumerate}[leftmargin=*]
    \item \textbf{Taxonomic edges} $(c_{\text{parent}},c_{\text{child}})$ encode the ``is–a'' relation of the hierarchy tree.
    \item \textbf{Tag--category edges} $(t,c)$ for every tag $t$ assigned to a leaf category $c$.
    \item \textbf{Model--tag edges} $(m,t)$ whenever the model $m$ carries the tag $t$.
    \item \textbf{Co-occurrence edges} $(t_i,t_j)$ with weight $w_{ij}$ proportional to the point-wise mutual information or normalized co-frequency of $t_i$ and $t_j$ across models.
\end{enumerate}

\vspace{0.1cm}
In the future, we aim to equip each node with learned feature vectors, \textit{i.e.,} text embeddings generated using their captions passed through a foundational model like GPT\cite{achiam2023gpt}. The resulting structure is a richly typed knowledge graph that captures both symbolic semantics (``V-shaped structural bracket'') and descriptions.

\vspace{-0.2cm}
\subsection{Graph Mining and Retrieval Mechanism}\label{supplsubsec:Graph_Mining_Retrieval}
\vspace{-0.1cm}
Given this representation, both discovery and retrieval reduce to structured navigation over the graph.
\vspace{0.15cm}
\noindent\textbf{Mining semantic and geometric patterns.}
We apply community detection and frequent subgraph mining to the tag-tag and model-tag subgraphs to uncover recurrent design motifs.  For example, high-weight cliques connecting \emph{box shape}, \emph{rectangular pocket}, and \emph{through hole} correspond to machining templates frequently seen in prismatic parts, while another community might link \emph{axial gear}, \emph{shaft}, and \emph{keyway}.  At higher levels, modularity-based clustering over the category graph yields the thematic groupings visualized in the first image, where related subtrees (e.g., gear systems, bent profiles, and multi-angled structures) are summarized as coherent "families" of shapes. These mined structures directly inform both dataset balancing and the design of tag suggestions for new models.

\begin{figure*}[t]
  \centering
  \includegraphics[width=0.99\linewidth]{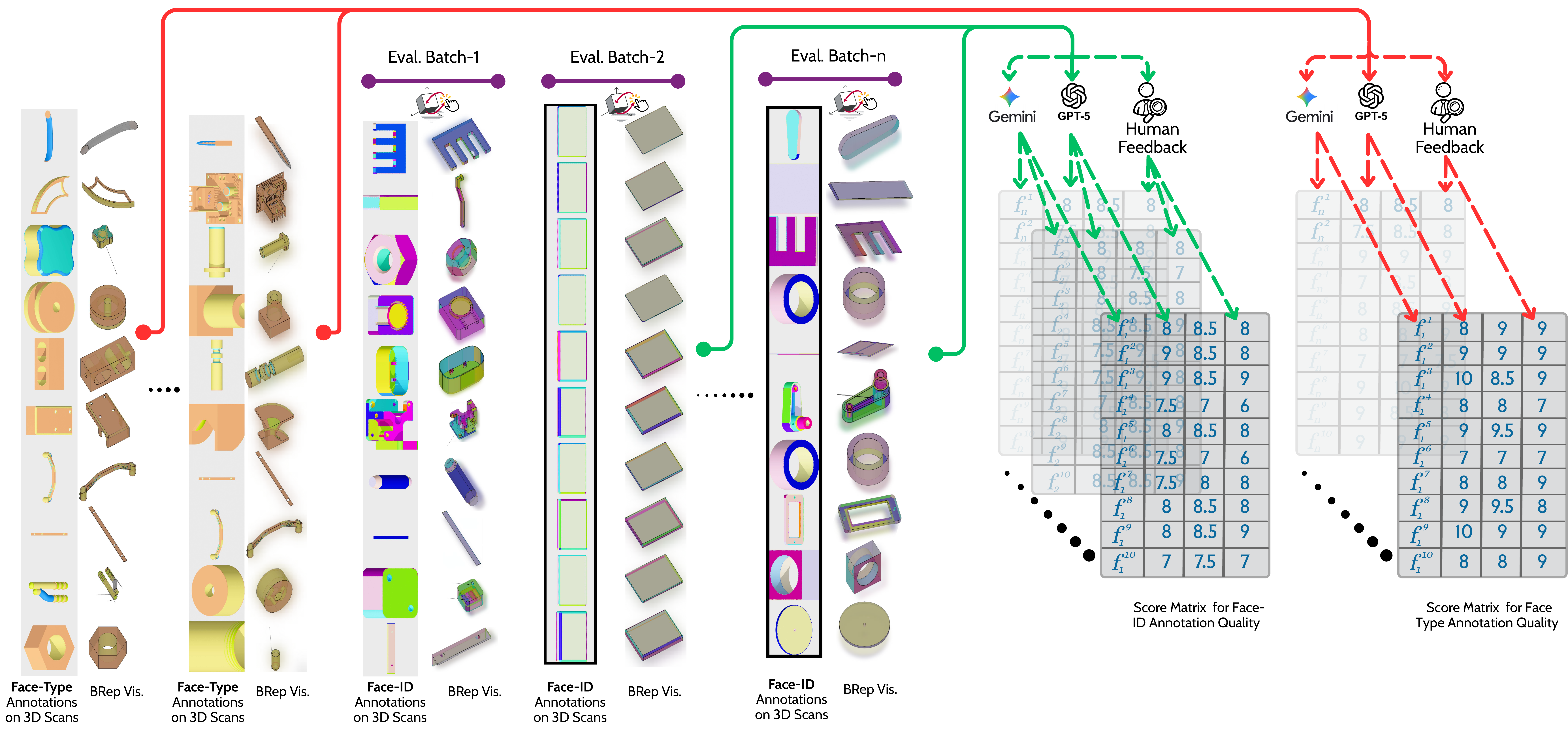}
   \caption{This figure visually describes how evaluation and scoring mechanisms from GPT-5, Gemini, and the Human feedback-based evaluation process for different types of annotations on 10K randomly selected samples.}
   \label{fig:SupplyQualityEval_Proc}
\end{figure*}
\vspace{-0.2cm}
\section{GPT-5, Gemini, and Human Evaluation.}\label{supply:sec:QualityEval_Proc}
\vspace{-0.15cm}
In continuation of \textit{Section 5} of the main matter, the \cref{fig:SupplyQualityEval_Proc} describes how a multi-stage workflow was designed to measure the quality of \textbf{\textit{Face-Type, Face-ID, Boundary-Type,}} and \textbf{\textit{Boundary-ID}} annotations on both 3D scans and their corresponding B-Rep visualizations. As illustrated in the evaluation diagram, the randomly selected 10K samples are first divided into multiple “Eval Batches”, each containing 10 CAD models. For every batch, the 3D scans were color-mapped using the BRep entity feature (see the color map in the teaser figure of the main matter). Thereafter, a corresponding BRep model is also visualized using OpenCASCADE, where the geometric entities are color-mapped to serve as ground truth. Automatic quick snapshots are captured when the CAD models and 3D Scans are collectively rotated to other views. These stacked images are passed to Gemini, GPT-5 to collect the annotation quality scores. The human evaluators also submit their scores in parallel. These scores are aggregated into structured matrices—one for each type of BRep entity labels on the right side of \cref{fig:SupplyQualityEval_Proc}. To avoid redundancy, the other similar matrices for sketches and boundaries are not illustrated. However, we have conducted a similar evaluation for all the boundary, corner, and face-related BRep entities. 
\vspace{-0.15cm}
\subsection{Chunk-wise Evaluation of Model}\label{supply:subsec:Unroll_100_Chunks}
\begin{figure*}[t]
  \centering
  \includegraphics[width=0.9\linewidth, keepaspectratio, clip]{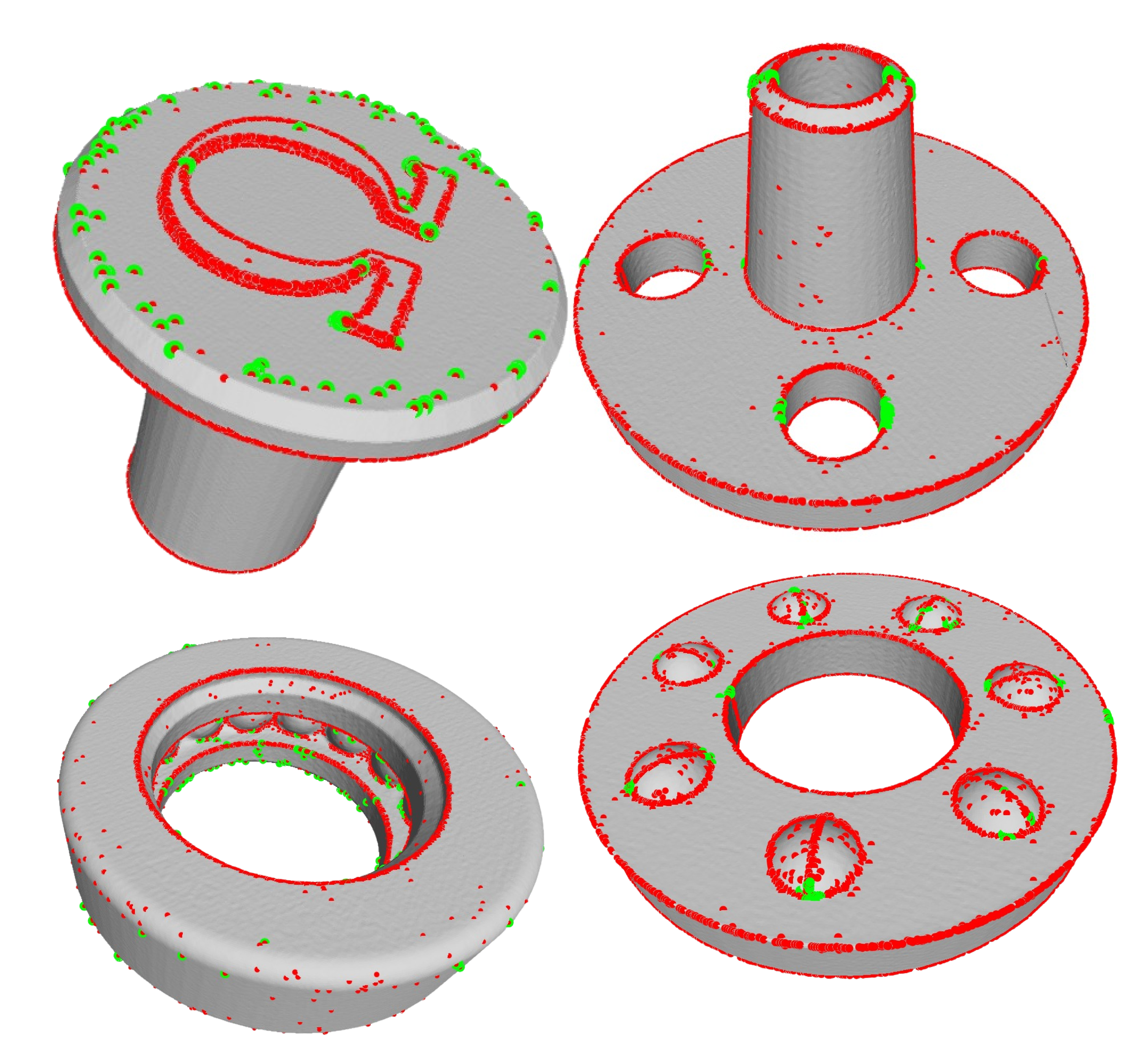}
  \caption{Boundary and Corner detection on randomly selected real-world 3D scan data from the Grab-CAD repository.}
  \label{fig:randomBJ}
\end{figure*}

\vspace{-0.15cm}
We have also unrolled the chunk-wise (0--100) recall and precision measurements for the boundary and junction tasks in \cref{supply:Table:Boundary_Junction_Eval}. The green colored region, \textit{i.e.,} chunks 70–99, contains 3D scans of a very high-resolution (16x upsampled version of ABC meshes). Therefore, the recall-precision for those chunks is consistently strong. These unrolled results complement the average metrics reported in the main paper. The \cref{fig:randomBJ} illustrated very high-quality boundary and corner detection on a randomly downloaded 3D scan file from the online GrabCAD (\url{https://grabcad.com/library}) library  
\setkeys{Gin}{width=\linewidth, height=\textheight, keepaspectratio}
\begin{figure*}[!t]
  \centering
  \includegraphics[width=0.8\linewidth]{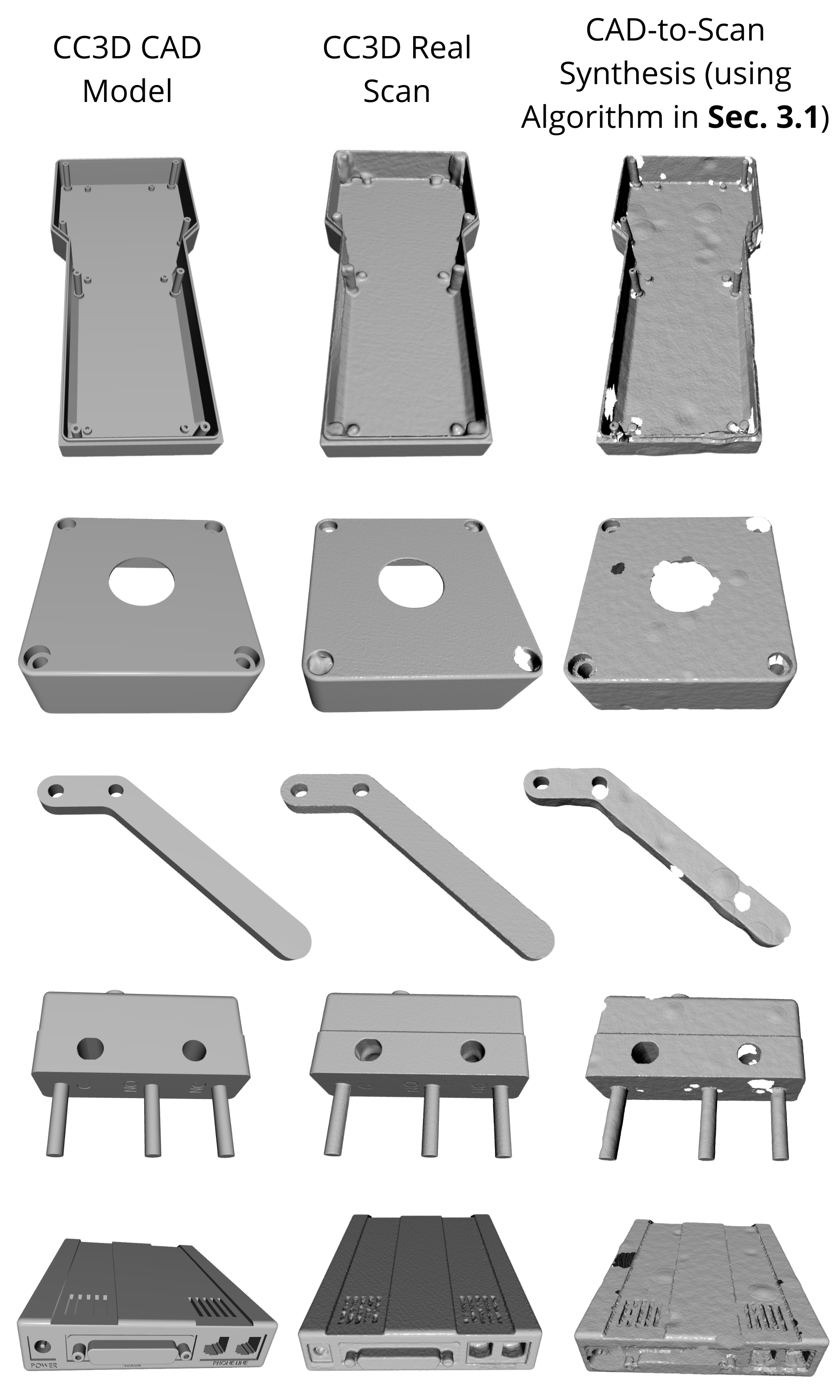}
   \caption{Visual comparison between CC3D \cite{cc3d} CAD models, its real scan pairs, and \textbf{synthesized scans from CAD (ours)}}
   \label{fig:CC3DScanVsSynthesis}
\end{figure*}
\vspace{-0.2cm}
\section{More Examples of Caption and Tags}\label{supplysubsec:More_results_Caption}
\begin{figure*}[t]
  \centering
  \includegraphics[width=0.99\linewidth]{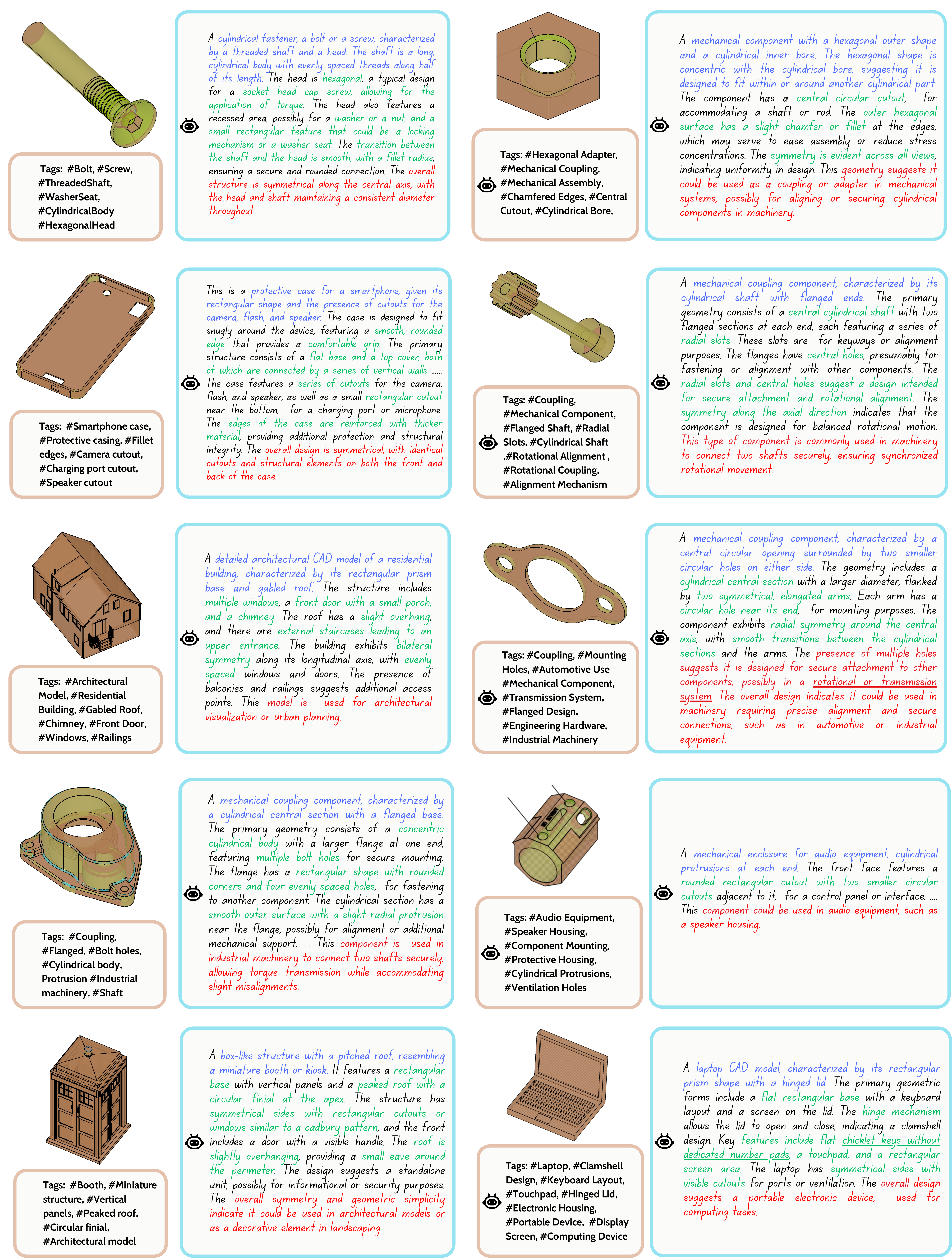}
   \caption{Visual elaboration of the textual captions on more complex CAD models.}
   \label{fig:SupplyCaptionTags_1}
\end{figure*}

\begin{figure*}[t]
  \centering
  \includegraphics[width=0.99\linewidth]{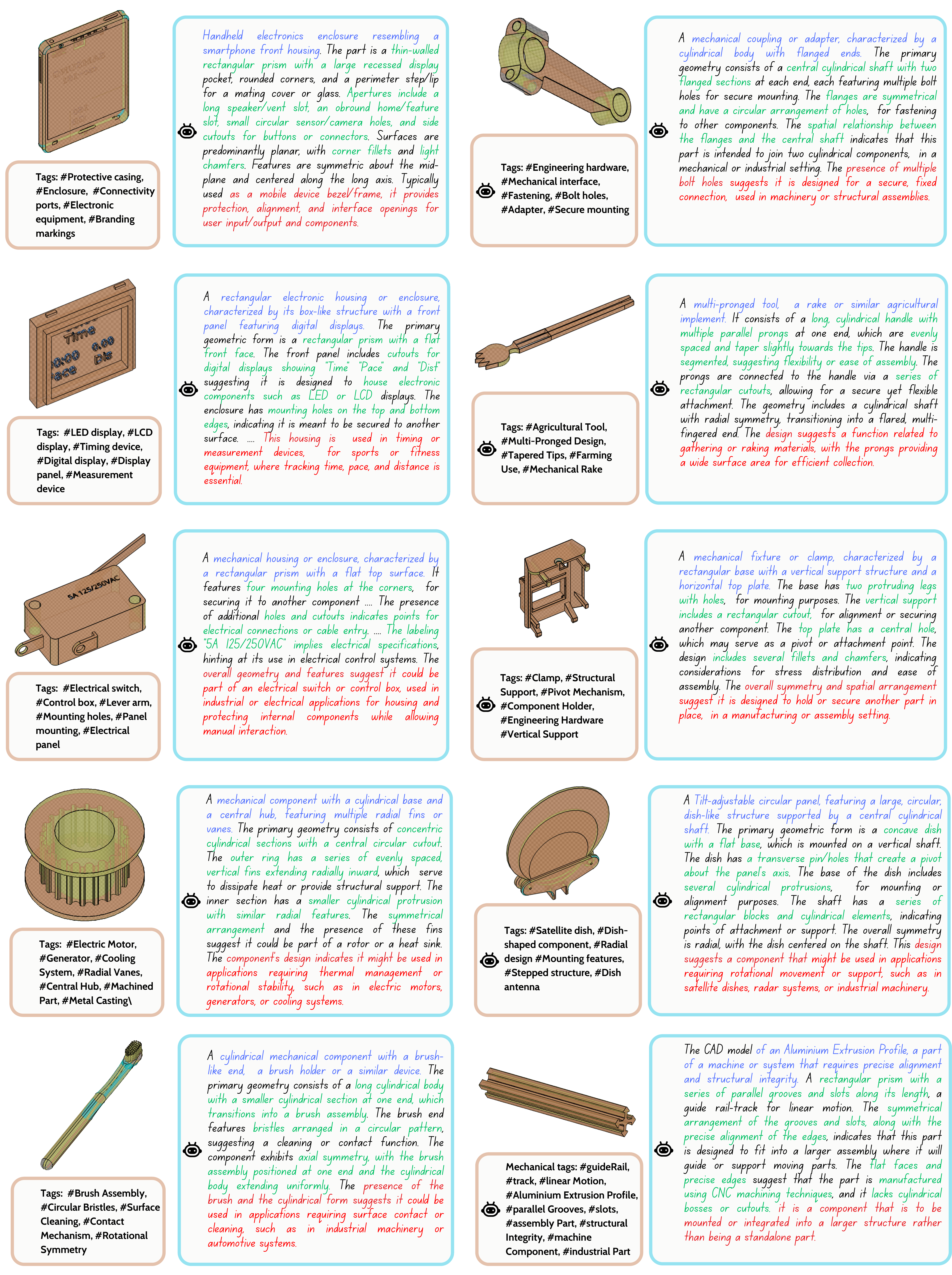}
   \caption{Visual elaboration of the textual captions on more complex CAD models}
   \label{fig:SupplyCaptionTags_2}
\end{figure*}

\begin{figure*}[t]
  \centering
  \includegraphics[width=0.99\linewidth]{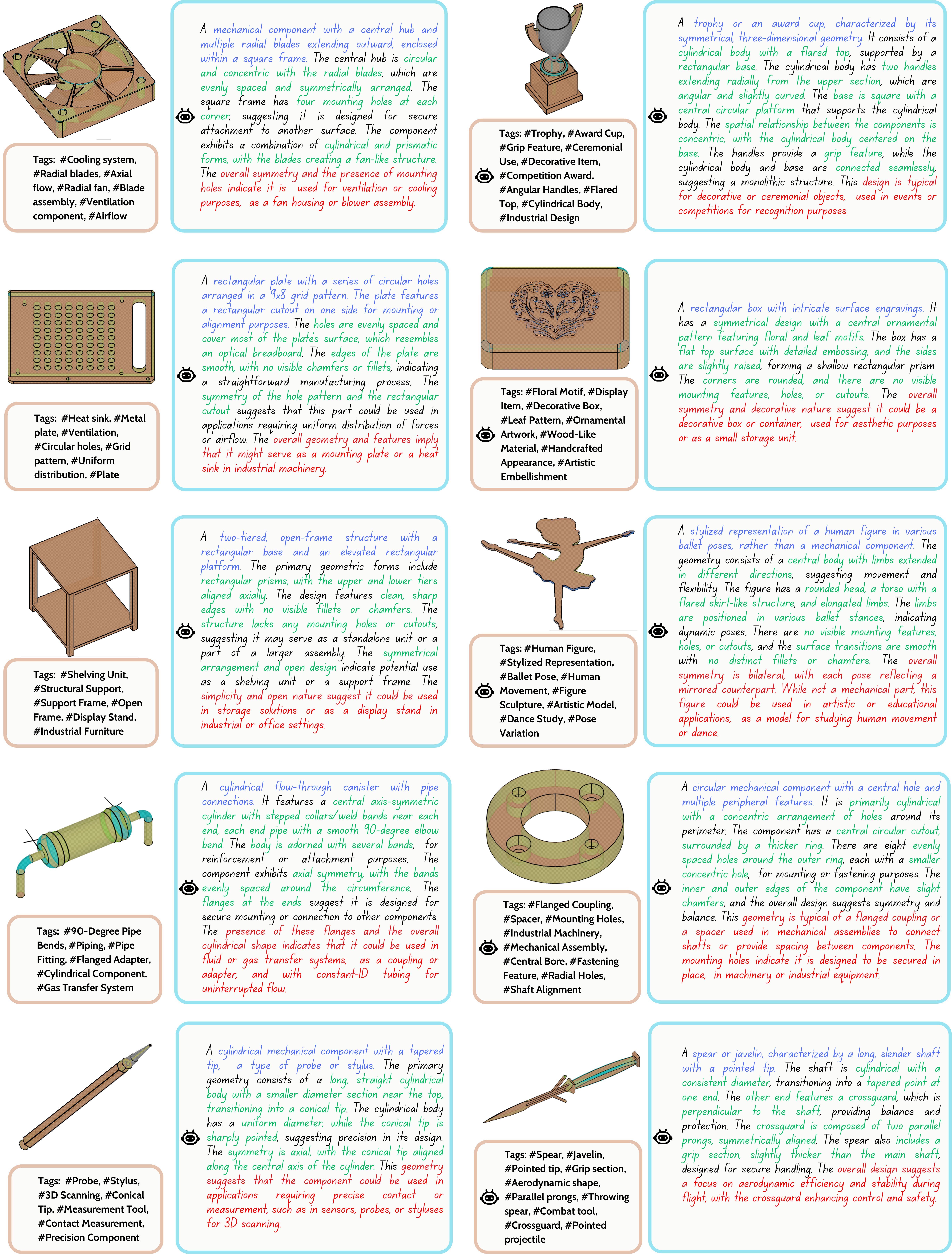}
   \caption{Visual elaboration of the textual captions on more complex CAD models}
   \label{fig:SupplyCaptionTags_3}
\end{figure*}

\vspace{-0.1cm}
\cref{fig:SupplyCaptionTags_1,fig:SupplyCaptionTags_2,fig:SupplyCaptionTags_3} illustrates many examples of CAD captions and tags fetched randomly from our \textbf{A2Z}. The example captions and tags reveal a \textcolor{blue}{\textit{geometry-first}} reasoning style: each description starts by carefully characterizing observable shapes (cylinders, plates, frames, blades) and spatial relations (radial, concentric, open-frame), and only then moves to \emph{functional hypotheses} such as cooling, support, or grip.  Across both mechanical and decorative objects, the text consistently integrates \textcolor{teal}{\textit{local features}} (holes, flanges, chamfers, cut-outs) with \textcolor{teal}{\textit{global structure}} (symmetry, layering, pose) to justify why a part might behave as a fan housing, a probe, a bracket-like stand, or a stylized human figure.  The sets of tags act as compact \textcolor{magenta}{\textit{multi-axis descriptors}}, jointly encoding category (e.g., component vs. figurine), function (cooling, fastening, display), dominant geometry (cylindrical body, radial blades, open frame), and likely application domain (industrial machinery, ventilation, artistic display).  Together, these captions and tags show how language can remain tightly grounded in CAD evidence while still offering informative, searchable abstractions that bridge \emph{shape}, \emph{function}, and \emph{use context}.
\vspace{-0.15cm}
\section{More Examples of Electronic Enclosures}\label{supplysec:CADforElectronicEnclosures}
\begin{figure*}[t]
  \centering
\includegraphics[width=0.92\linewidth]{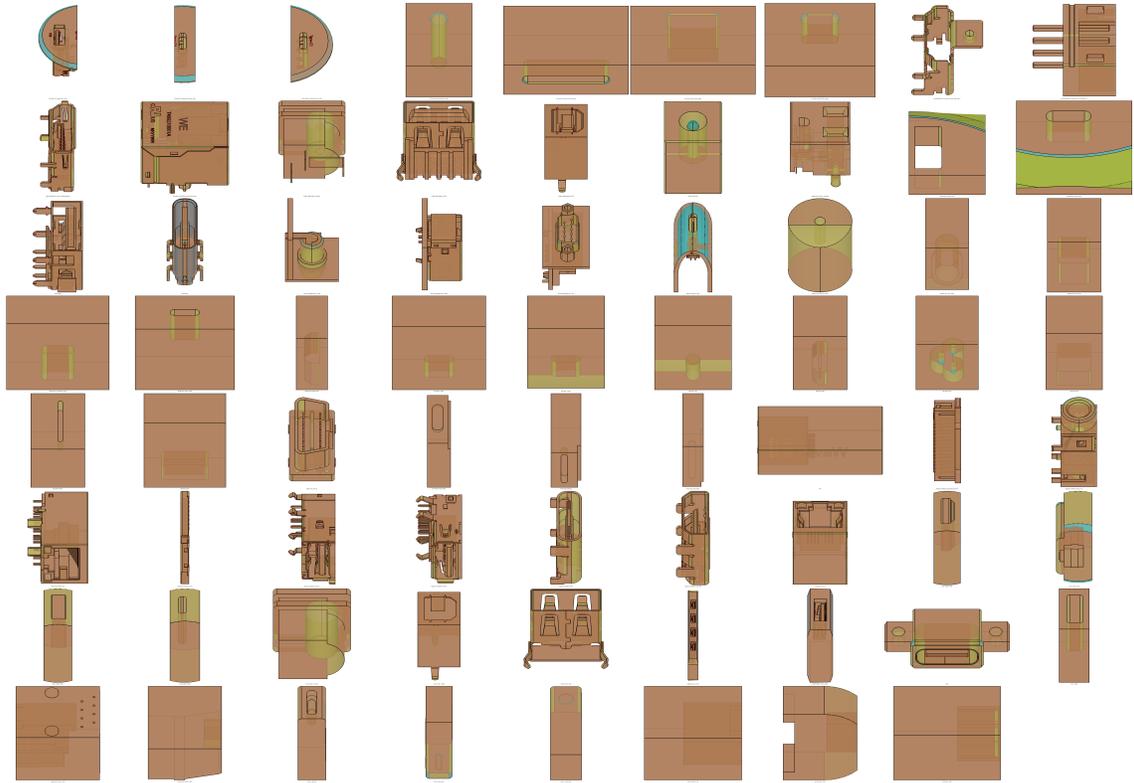}
   \caption{Visuals of some randomly selected CAD models of charging ports under \textbf{A2Z}. Color codes define face-Types.}
   \label{fig:supply_Ports_facetypes_Extra25K}
\end{figure*}

\begin{figure*}[t]
  \centering
  \includegraphics[width=0.92\linewidth]{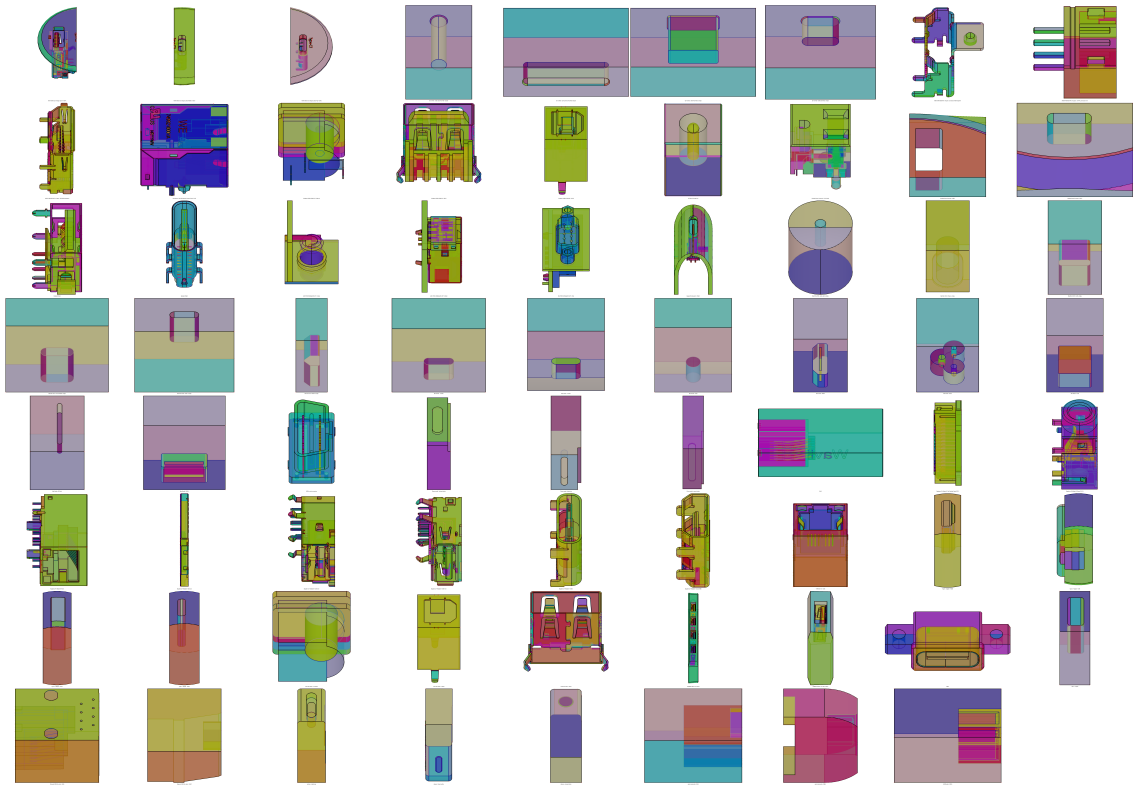}
   \caption{Visuals of same CAD models of charging ports under \textbf{A2Z} where color codes define face-IDs.}
   \label{fig:supply_Ports_faceIDs_Extra25K}
\end{figure*}

\begin{figure*}[t]
  \centering
  \includegraphics[width=0.85\linewidth]{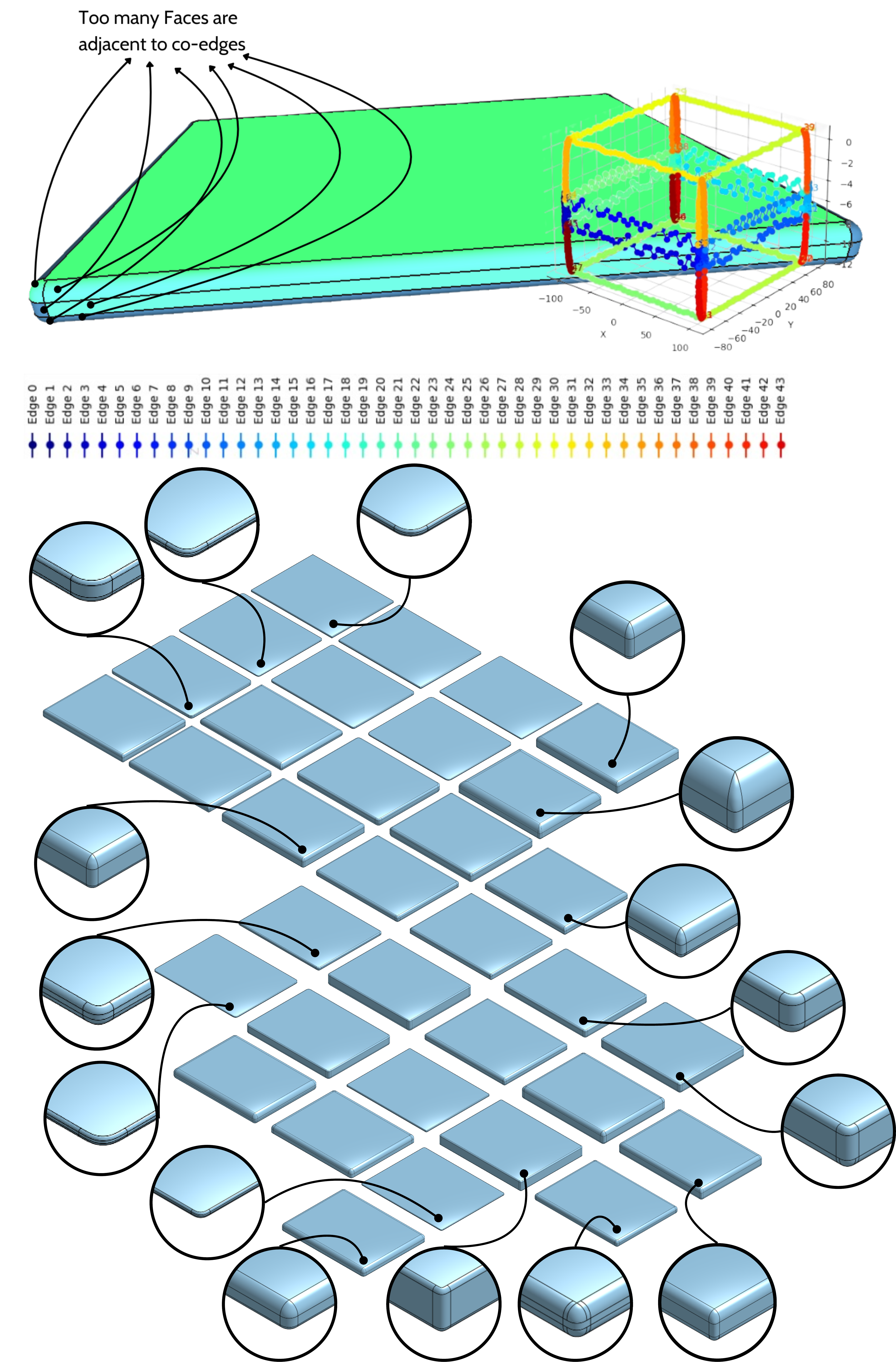}
   \caption{Visuals of some randomly selected samples of CAD models for Tablet type enclosures (from Approx. 20K Electronic enclosure models) with varying corner geometry and topology (\textit{at bottom}). A closer look at the BRep faces, edges, and vertices near the corner parts (\textit{on top}) shows how many faces are adjacent to each edge.}
   \label{fig:Tablets_Closures}
\end{figure*}

\vspace{-0.15cm}
The \cref{fig:supply_Ports_facetypes_Extra25K} and \cref{fig:supply_Ports_faceIDs_Extra25K} demonstrate samples that were randomly drawn from our newly added CAD models in \textbf{A2Z}, consisting of charging ports, usb-a/b/c type ports, and other socket-type enclosures. The present CAD market for electronic equipment is larger than that for mechanical engineering. The newly added CAD models and results from our foundational model of boundary and junction detection will be highly effective in standardizing the BRep entities of such complex CAD models. 
Unlike the charging port design data, where the complexity of CAD models is greater in terms of the topology of the folding faces/boundaries, the \cref{fig:Tablets_Closures} illustrates a different challenge regarding the geometric continuity of the BRep faces/boundaries around the corner region. This type of data annotation and our foundational model for parsing the BRep entities are close to industrial settings for accurate CAD reconstruction.

\vspace{-0.15cm}
\section{Examples of Annotations and Predictions}\label{suppl:subsec:Scan_Skecth_Labels}
\begin{figure*}[t]
  \centering
  \includegraphics[width=0.9\linewidth]{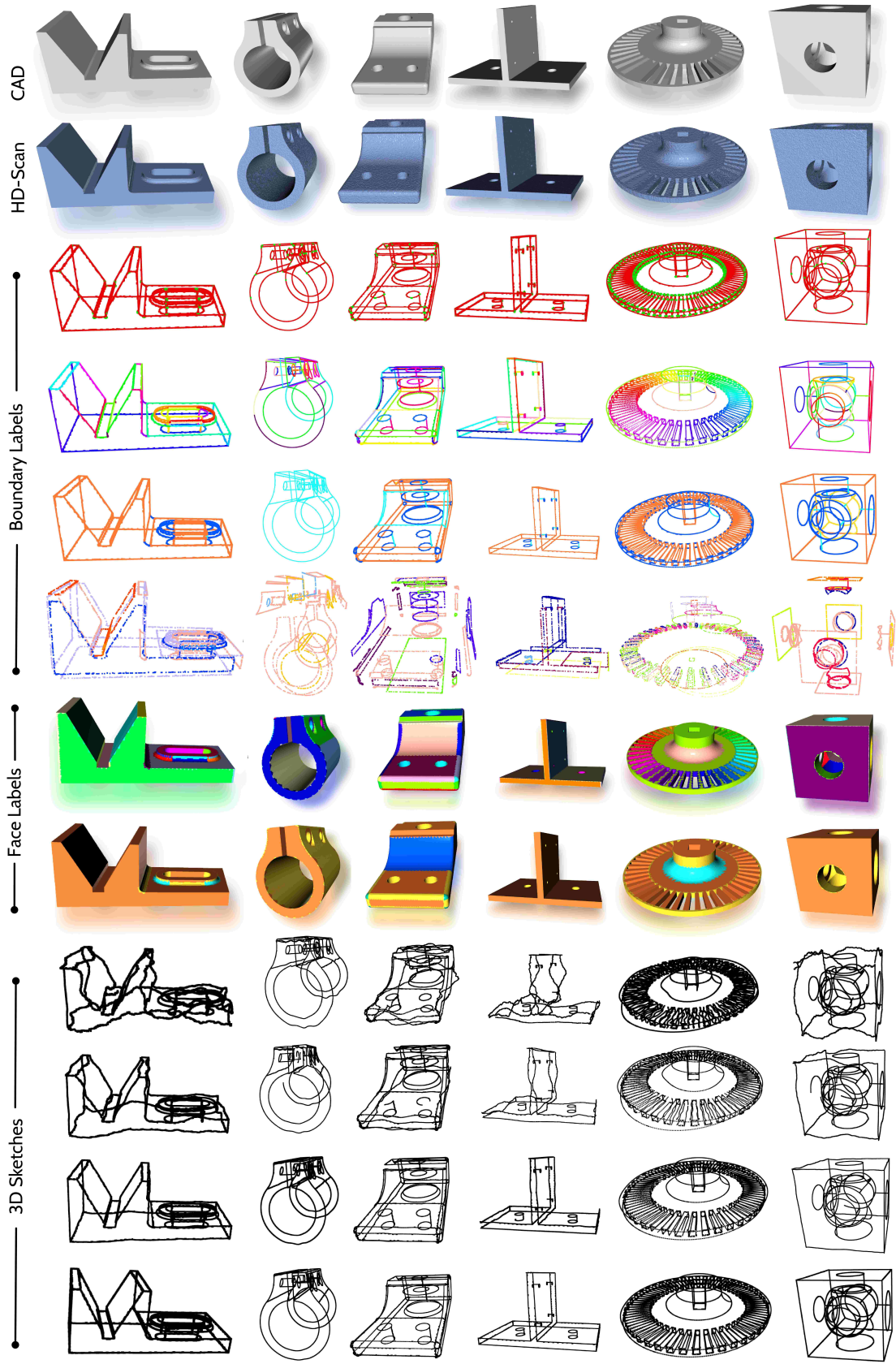}
   \caption{More visual examples of BRep Annotations for reverse engineering from 3D scans/sketches.}
   \label{fig:Supply3DAnnotations}
\end{figure*}

\vspace{-0.15cm}
In \cref{fig:Supply3DAnnotations} we add more examples of our rich multi-modal annotations on 3D scans and sketches. The color coding is shown in Fig.\,3 of the main matter. 
\begin{table*}[t]
\centering
\begin{minipage}[t]{0.49\textwidth}
\centering
\begin{tabular}{rcccc}
\toprule
\textbf{Chunk} & \multicolumn{2}{c}{\textbf{Boundary}} & \multicolumn{2}{c}{\textbf{Junction}} \\
\cmidrule(lr){2-3}\cmidrule(lr){4-5}
& \textbf{Recall} & \textbf{Precision} & \textbf{Recall} & \textbf{Precision} \\
\midrule
0  & 0.972 & 0.885 & 0.690 & 0.740 \\
1  & 0.972 & 0.905 & 0.710 & 0.774 \\
2  & 0.980 & 0.883 & 0.681 & 0.703 \\
3  & 0.977 & 0.883 & 0.688 & 0.739 \\
4  & 0.964 & 0.841 & 0.698 & 0.701 \\
5  & 0.977 & 0.877 & 0.688 & 0.765 \\
6  & 0.972 & 0.889 & 0.673 & 0.764 \\
7  & 0.958 & 0.851 & 0.690 & 0.791 \\
8  & 0.961 & 0.879 & 0.682 & 0.708 \\
9  & 0.972 & 0.877 & 0.682 & 0.731 \\
10 & 0.982 & 0.878 & 0.692 & 0.780 \\
11 & 0.973 & 0.878 & 0.689 & 0.710 \\
12 & 0.965 & 0.868 & 0.698 & 0.711 \\
13 & 0.980 & 0.895 & 0.678 & 0.737 \\
14 & 0.970 & 0.900 & 0.696 & 0.736 \\
15 & 0.974 & 0.890 & 0.683 & 0.747 \\
16 & 0.939 & 0.876 & 0.703 & 0.740 \\
17 & 0.974 & 0.864 & 0.682 & 0.680 \\
18 & 0.958 & 0.874 & 0.674 & 0.778 \\
19 & 0.973 & 0.886 & 0.679 & 0.745 \\
20 & 0.959 & 0.870 & 0.702 & 0.755 \\
21 & 0.971 & 0.881 & 0.688 & 0.707 \\
22 & 0.979 & 0.886 & 0.680 & 0.765 \\
23 & 0.976 & 0.891 & 0.678 & 0.747 \\
24 & 0.979 & 0.898 & 0.683 & 0.771 \\
25 & 0.978 & 0.876 & 0.681 & 0.748 \\
26 & 0.963 & 0.880 & 0.688 & 0.736 \\
27 & 0.963 & 0.857 & 0.695 & 0.739 \\
28 & 0.956 & 0.870 & 0.695 & 0.692 \\
29 & 0.970 & 0.887 & 0.681 & 0.763 \\
30 & 0.967 & 0.891 & 0.695 & 0.799 \\
31 & 0.970 & 0.885 & 0.651 & 0.744 \\
32 & 0.972 & 0.896 & 0.658 & 0.777 \\
33 & 0.964 & 0.848 & 0.656 & 0.715 \\
34 & 0.979 & 0.901 & 0.658 & 0.714 \\
35 & 0.965 & 0.873 & 0.658 & 0.726 \\
36 & 0.980 & 0.895 & 0.651 & 0.714 \\
37 & 0.958 & 0.852 & 0.648 & 0.794 \\
38 & 0.968 & 0.871 & 0.621 & 0.759 \\
39 & 0.975 & 0.872 & 0.674 & 0.724 \\
40 & 0.978 & 0.903 & 0.660 & 0.731 \\
41 & 0.974 & 0.866 & 0.662 & 0.756 \\
42 & 0.967 & 0.886 & 0.659 & 0.731 \\
43 & 0.964 & 0.874 & 0.656 & 0.700 \\
44 & 0.948 & 0.861 & 0.653 & 0.751 \\
45 & 0.972 & 0.872 & 0.659 & 0.727 \\
46 & 0.966 & 0.888 & 0.673 & 0.766 \\
47 & 0.976 & 0.894 & 0.638 & 0.726 \\
48 & 0.979 & 0.881 & 0.639 & 0.741 \\
49 & 0.961 & 0.864 & 0.635 & 0.721 \\
\bottomrule
\end{tabular}
\end{minipage}\hfill
\begin{minipage}[t]{0.49\textwidth}
\centering
\begin{tabular}{rcccc}
\toprule
\textbf{Chunk} & \multicolumn{2}{c}{\textbf{Boundary}} & \multicolumn{2}{c}{\textbf{Junction}} \\
\cmidrule(lr){2-3}\cmidrule(lr){4-5}
& \textbf{Recall} & \textbf{Precision} & \textbf{Recall} & \textbf{Precision} \\
\midrule
50 & 0.974 & 0.888 & 0.642 & 0.728 \\
51 & 0.980 & 0.900 & 0.670 & 0.723 \\
52 & 0.967 & 0.885 & 0.665 & 0.739 \\
53 & 0.971 & 0.882 & 0.675 & 0.726 \\
54 & 0.965 & 0.876 & 0.639 & 0.734 \\
55 & 0.963 & 0.895 & 0.668 & 0.759 \\
56 & 0.950 & 0.884 & 0.629 & 0.726 \\
57 & 0.966 & 0.886 & 0.686 & 0.749 \\
58 & 0.970 & 0.888 & 0.673 & 0.743 \\
59 & 0.970 & 0.885 & 0.645 & 0.748 \\
60 & 0.965 & 0.885 & 0.641 & 0.756 \\
61 & 0.975 & 0.876 & 0.657 & 0.742 \\
62 & 0.979 & 0.887 & 0.668 & 0.720 \\
63 & 0.972 & 0.882 & 0.674 & 0.743 \\
64 & 0.967 & 0.867 & 0.641 & 0.676 \\
65 & 0.967 & 0.911 & 0.682 & 0.760 \\
66 & 0.965 & 0.885 & 0.640 & 0.716 \\
67 & 0.963 & 0.878 & 0.637 & 0.757 \\
68 & 0.978 & 0.888 & 0.684 & 0.748 \\
69 & 0.974 & 0.894 & 0.657 & 0.763 \\
70 & \good{0.981} & \good{0.909} & \good{0.781} & \good{0.897} \\
71 & \good{0.978} & \good{0.892} & \good{0.775} & \good{0.917} \\
72 & \good{0.980} & \good{0.890} & \good{0.777} & \good{0.931} \\
73 & \good{0.987} & \good{0.909} & \good{0.782} & \good{0.928} \\
74 & \good{0.983} & \good{0.900} & \good{0.776} & \good{0.902} \\
75 & \good{0.983} & \good{0.899} & \good{0.772} & \good{0.893} \\
76 & \good{0.985} & \good{0.896} & \good{0.777} & \good{0.938} \\
77 & \good{0.986} & \good{0.884} & \good{0.783} & \good{0.909} \\
78 & \good{0.986} & \good{0.913} & \good{0.775} & \good{0.873} \\
79 & \good{0.980} & \good{0.895} & \good{0.777} & \good{0.941} \\
80 & \good{0.985} & \good{0.900} & \good{0.784} & \good{0.940} \\
81 & \good{0.981} & \good{0.893} & \good{0.775} & \good{0.917} \\
82 & \good{0.981} & \good{0.896} & \good{0.741} & \good{0.917} \\
83 & \good{0.987} & \good{0.901} & \good{0.782} & \good{0.932} \\
84 & \good{0.984} & \good{0.905} & \good{0.779} & \good{0.844} \\
85 & \good{0.985} & \good{0.917} & \good{0.767} & \good{0.869} \\
86 & \good{0.981} & \good{0.883} & \good{0.744} & \good{0.929} \\
87 & \good{0.983} & \good{0.895} & \good{0.740} & \good{0.912} \\
88 & \good{0.984} & \good{0.905} & \good{0.742} & \good{0.895} \\
89 & \good{0.986} & \good{0.895} & \good{0.740} & \good{0.872} \\
90 & \good{0.977} & \good{0.894} & \good{0.748} & \good{0.913} \\
91 & \good{0.981} & \good{0.896} & \good{0.748} & \good{0.931} \\
92 & \good{0.984} & \good{0.885} & \good{0.752} & \good{0.915} \\
93 & \good{0.984} & \good{0.891} & \good{0.742} & \good{0.899} \\
94 & \good{0.983} & \good{0.902} & \good{0.742} & \good{0.934} \\
95 & \good{0.986} & \good{0.903} & \good{0.741} & \good{0.915} \\
96 & \good{0.982} & \good{0.877} & \good{0.745} & \good{0.912} \\
97 & \good{0.983} & \good{0.886} & \good{0.778} & \good{0.920} \\
98 & \good{0.982} & \good{0.884} & \good{0.779} & \good{0.924} \\
99 & \good{0.977} & \good{0.870} & \good{0.781} & \good{0.893} \\
\bottomrule
\end{tabular}
\end{minipage}
\caption{Precision-recall measures for all 100 chunks of \textbf{A2Z} using our foundation model.}
\label{supply:Table:Boundary_Junction_Eval}
\end{table*}


\end{document}